\documentclass[10pt,twocolumn,letterpaper]{article}

\usepackage{cvpr}              

\usepackage{times}

\usepackage[utf8]{inputenc} 
\usepackage[T1]{fontenc}    
\usepackage{url}            
\usepackage{booktabs}       
\usepackage{amsfonts}       
\usepackage{nicefrac}       
\usepackage{microtype}      
\usepackage{xcolor}         

\usepackage{graphicx}
\usepackage{amsmath,amssymb}
\usepackage{verbatim}
\usepackage{multirow}
\usepackage{marvosym}
\usepackage{makecell}
\usepackage{bm}
\usepackage{subfloat}
\usepackage{wrapfig}
\usepackage{caption}

\usepackage[accsupp]{axessibility}  

\newcommand{\R}{\ensuremath{\mathbb{R}}}

\newcommand{\mypar}[1]{{\bf #1.}}

\def\x{\mathbf{x}}

\def\X{\mathbf{X}}

\usepackage[pagebackref,breaklinks,colorlinks]{hyperref}

\usepackage[capitalize]{cleveref}
\crefname{section}{Sec.}{Secs.}
\Crefname{section}{Section}{Sections}
\Crefname{table}{Table}{Tables}
\crefname{table}{Tab.}{Tabs.}

\def\cvprPaperID{552} 
\def\confName{CVPR}
\def\confYear{2022}

\begin{document}

\title{GroupNet: Multiscale Hypergraph Neural Networks for \\ Trajectory Prediction with Relational Reasoning}

\author{
Chenxin Xu\textsuperscript{1}, Maosen Li\textsuperscript{1}, Zhenyang Ni\textsuperscript{1}, Ya Zhang\textsuperscript{1,2}, Siheng Chen\textsuperscript{1,2\footnotemark[1]}
\\\textsuperscript{1}Shanghai Jiao Tong University,  \textsuperscript{2}Shanghai AI Laboratory
\\
{\tt\small \{xcxwakaka,maosen\_li,0107nzy,ya\_zhang,sihengc\}@sjtu.edu.cn}
}

\maketitle

\renewcommand{\thefootnote}{\fnsymbol{footnote}}
\footnotetext[1]{Corresponding author.}\footnotetext{Code is available at: \url{https://github.com/MediaBrain-SJTU/GroupNet}}

\begin{abstract}
Demystifying the interactions among multiple agents from their past trajectories is fundamental to precise and interpretable trajectory prediction. However, previous works only consider pair-wise interactions with limited relational reasoning. To promote more comprehensive interaction modeling for relational reasoning, we propose GroupNet, a multiscale hypergraph neural network, which is novel in terms of both interaction capturing and representation learning. From the aspect of interaction capturing, we propose a trainable multiscale hypergraph to capture both pair-wise and group-wise interactions at multiple group sizes. From the aspect of interaction representation learning, we propose a three-element format that can be learnt end-to-end and explicitly reason some relational factors including the interaction strength and category. We apply GroupNet into both CVAE-based prediction system and previous state-of-the-art prediction systems for predicting socially plausible trajectories with relational reasoning. To validate the ability of relational reasoning, we experiment with synthetic physics simulations to reflect the ability to capture group behaviors, reason interaction strength and interaction category. To validate the effectiveness of prediction, we conduct extensive experiments on three real-world trajectory prediction datasets, including NBA, SDD and ETH-UCY; and we show that with GroupNet, the CVAE-based prediction system outperforms state-of-the-art methods. We also show that adding GroupNet will further improve the performance of previous state-of-the-art prediction systems. 
\end{abstract}

\vspace{-1mm}
\section{Introduction}
\vspace{-1mm}
\label{sec:intro}

Multi-agent trajectory prediction aims to predict future trajectories of agents conditioned on past movements. This task is not only critical to numerous real-world applications, such as autonomous driving and human behavior understanding, but also serves as a foundation to bridge the knowledge of the past and the action for the future. At least three factors would affect each agent's dynamics: ego momentum, instantaneous intent, and social influence from the other agents. The first factor has been well studied {\cite{sutskever2014sequence}}; the second factor is unpredictable; and the third factor is an emerging research topic and the focus of this work. To demystify the social influence, we need to model and reason the interactions among agents based on their past spatio-temporal states, potentially leading to precise and interpretable trajectory prediction.

\begin{figure}[t] 
\centering
\includegraphics[width=0.46\textwidth]{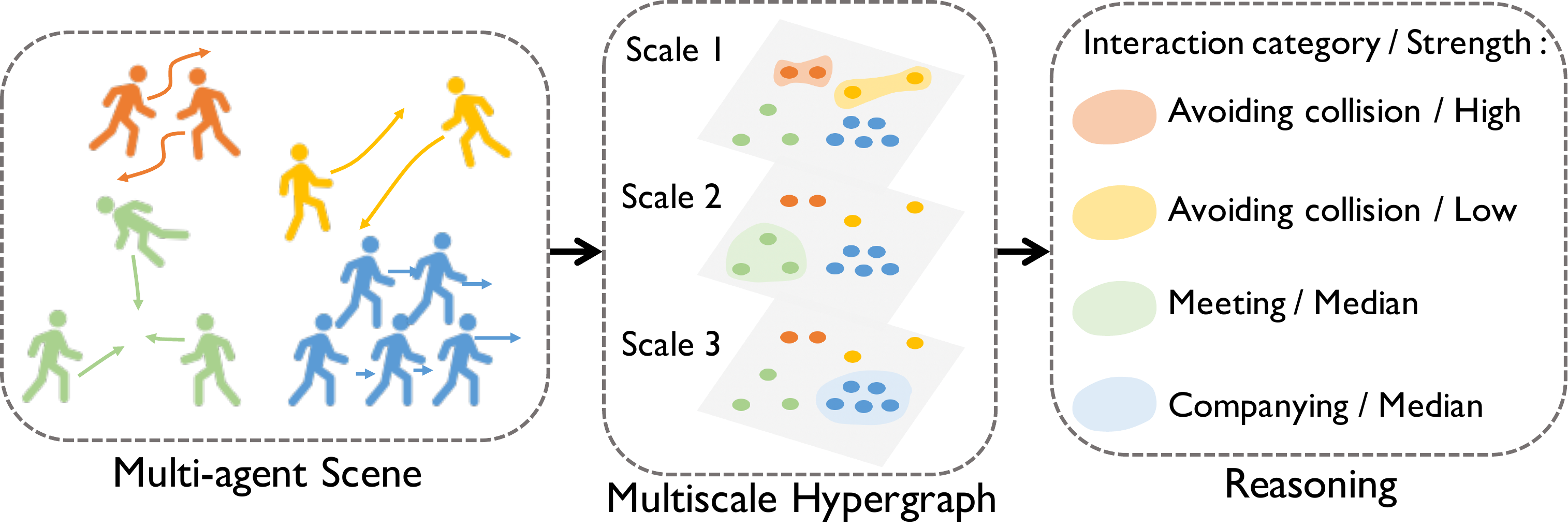}
\vspace{-3mm}
\caption{\small GroupNet captures both pair-wise and group-wise interactions among multiple agents from their trajectories  and infers the interaction category and strength for each interaction group.}
\label{fig:jewel}
\vspace{-17pt}
\end{figure}

Many works have tried to model the interactions through social operations \cite{alahi2016social,gupta2018social,xu2018encoding}, graph-based modeling \cite{kosaraju2019social,huang2019stgat,hu2020collaborative,li2020evolvegraph} and attention mechanism \cite{vemula2018social,sadeghian2019sophie,mangalam2020not}. Although these methods have greatly improved the empirical performances of trajectory prediction, they underestimate the significance of relational reasoning. Some previous works, such as NRI \cite{kipf2018neural} and Evolvegraph \cite{li2020evolvegraph}, took a step forward towards explicit relational reasoning; however, they still have limitation from two aspects. First, only pair-wise interactions are modeled. In many scenarios, such as on a basketball court, a group of players execute a defensive strategy cooperatively; in the ocean, a school of fish coordinate their movements to evade predators, reflecting collective behaviors. While this kind of group-wise interaction is common, they have rarely been modeled. Second, only interaction categories are explicitly reasoned by the current methods, while the intensity level of the agent interaction cannot be reflected.

To further promote more comprehensive relational reasoning in trajectory prediction, this work puts efforts on two aspects: interaction capturing and interaction representation learning. To capture more interactions, we propose a multiscale hypergraph, which consists of a series of hypergraphs to model group-wise interactions with multiple group sizes. Instead of using handcrafted designs, we learn such a multiscale hypergraph topology in a data-driven manner. To learn the interaction embedding, we propose a three-element representation format: the neural interaction strength, the neural interaction category and the per-category function, which can reflect the interaction strength and category in an interactive group. Based on neural message passing over the multiscale hypergraph, we merge this three-element interaction embedding in the representation learning process. 

Overall, we integrate these two designs and propose GroupNet, a multiscale hypergraph neural network, to capture social interactions for better trajectory prediction. The proposed GroupNet has two significant advantages: 1) \emph{strong ability of relational reasoning}. Through extensive synthetic physics experiments, we validate that GroupNet is able to capture group behaviors, infer interaction strength and interaction category during trajectory prediction without any relation supervision; and 2) \emph{strong ability of plug-and-play.} As a general and comprehensive social modeling module, GroupNet can be easily plugged in a prediction framework and improve its previous performance. For example, based on a straightforward CVAE-based  framework, GroupNet  significantly improves the prediction performance, leading to competitive or leading results on three real-world prediction benchmarks. Furthermore, for those current state-of-the-art prediction models, NMMP \cite{hu2020collaborative} on NBA dataset, PECNet \cite{mangalam2020not} on SDD dataset and Trajectron++ \cite{salzmann2020trajectron++} on ETH-UCY datasets, GroupNet brings significant benefit and lifts their results to even better performances on each dataset. The main contributions of this paper are summarized as follow:

$\bullet$ We propose GroupNet, a novel multiscale hypergraph neural network to model complex social influences for better trajectory prediction. GroupNet can capture both pair-wise and group-wise interactions at various group sizes and reflect interaction strength and interaction category.

$\bullet$ We validate the relational-reasoning ability of GroupNet  via extensive synthetic simulations and show that GroupNet can capture group behaviors, interaction strength and interaction category in an unsupervised setting; see Section~\ref{sec:relation}.

$\bullet$ With GroupNet, a simple CVAE-based framework achieves competitive performances on three real-world prediction benchmarks. We also show many state-of-the-art prediction systems can be significantly improved by replacing their social modules with GroupNet; see Section~\ref{sec:results}.

\section{Related Works}
\vspace{-1mm}
\textbf{Trajectory prediction.} Traditional approaches use hand-crafted rules and energy potentials \cite{antonini2006discrete,lee2007trajectory,mehran2009abnormal,morris2009learning,wang2011trajectory,wang2008unsupervised}. For complex scenes, sequence-to-sequence models \cite{sutskever2014sequence,alahi2016social} are leveraged to encode trajectories individually. Recent works model the interactions to predict more social-plausible trajectories. \cite{chou2018predicting,deo2018convolutional,zhao2019multi,bansal2018chauffeurnet,casas2018intentnet} use a spatial-centric mechanism to represent trajectories and consider the spatial interaction. \cite{alahi2016social,gupta2018social,xu2018encoding} use a social mechanism to aggregate neighboring actors and broadcasts to each actor for neighboring information diffusion. Besides, attention mechanism \cite{vemula2018social,sadeghian2019sophie,mangalam2020not,yuan2021agentformer,tang2021collaborative} and transformer structure \cite{giuliari2021transformer,yu2020spatio} are used to capture agents' spatial and temporal dependencies. Graph-based methods \cite{wu2020connecting,kosaraju2019social,huang2019stgat,hu2020collaborative,li2020evolvegraph,mohamed2020social,li2021online,gao2020vectornet} are proposed to explicitly model the interaction between actors through non-grid structures. However, previous methods only focus on modeling the pair-wise interaction, ignoring the group behavior's influence on agents. In this work, we extend the graph-based mechanism from ordinary graphs to multiscale hypergraphs to capture group behaviors at different sizes, thereby modeling agent interactions more comprehensively.

\textbf{Relational reasoning.}
In multi-agent systems, only the trajectories of individual agents are available without knowledge of the underlying relations, thus it is challenging to explicitly infer agents' interaction and make relational reasoning.
Some works attempt to infer the interaction between agents. NRI \cite{kipf2018neural} infers a latent interaction graph via an autoencoder model. RAIN \cite{li2021rain} also infers an interaction graph using reinforcement learning to select edges. EvolveGraph \cite{li2020evolvegraph} and dNRI \cite{graber2020dynamic} consider a evolve mechanism and learn a dynamic interaction graph. Our work also tries on relational reasoning but the main differences are: i) we capture more interactions including both group-wise interaction and pair-wise interaction, comparing to only inferring pair-wise relationship on previous methods; and ii) we infer both interaction category and interaction strength while previous works only consider the interaction category.

\vspace{-1mm}
\section{Problem Formulation}
\vspace{-1mm}
The task of trajectory prediction with relational reasoning is to predict future trajectories of multiple agents and infer interaction relationship given past trajectories. Mathematically, let
$\mathbb{X}^{-} \in \R^{N \times T_{\rm p} \times 2}$ and $\mathbb{X}^{+} \in \R^{N \times T_{\rm f} \times 2}$ be the past and future trajectories of all the $N$ agents in the scene. Let $\X^{-}_i = \mathbb{X}^{-}_{i,:,:} = [\x^{-T_{\rm p}+1}_{i}, \x^{-T_{\rm p}+2}_{i},\cdots, \x^{0}_{i}] \in \R^{T_{\rm p} \times 2}$ and $\X^{+}_i = \mathbb{X}^{+}_{i,:,:} = [\x^{1}_{i}, \x^{2}_{i},\cdots, \x^{T_{\rm f}}_{i}] \in \R^{T_{\rm f} \times 2}$ be the past and future trajectories of the $i$th agent, where $\x_i^t \in \R^{2}$ is the 2D coordinate at time $t$. Our goal is to learn a prediction model $g(\cdot)$, so that the predicted future trajectories $\widehat{\mathbb{X}}^{+} = g( \mathbb{X}^{-})$ are as close to the ground-truth future trajectories $\mathbb{X}^{+}$ as possible. When a prediction model $g(\cdot)$ is capable to infer social influences among the agents without any ground-truth  supervision about interaction category or strength, this supplementary task is called~\emph{unsupervised relational reasoning}.

\begin{figure*}[t] 
\centering
\includegraphics[width=0.9\textwidth]{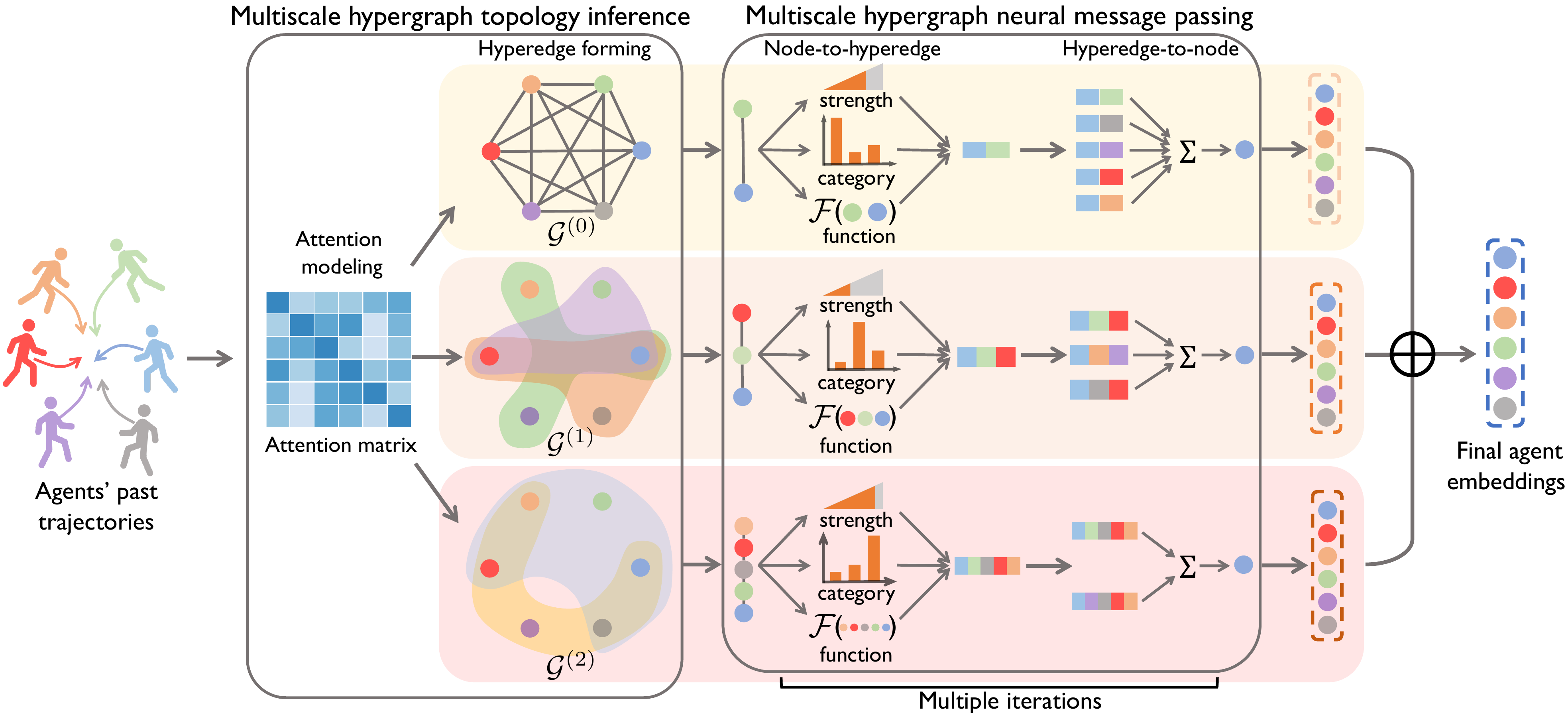}
\vspace{-5pt}
\caption{\small \textbf{The architecture of GroupNet.} GroupNet has two stages: multiscale hypergraph topology inference and multiscale hypergraph neural message passing, where we infer the hypergraph topologies and capture the agent patterns with their interaction, respectively. In multiscale hypergraph neural message passing, we model the interaction as three elements to reflect physical intuition and interpretation.
}
\label{fig:ms-nmp}
\vspace{-18pt}
\end{figure*}

\section{GroupNet}
\vspace{-1mm}
The core of GroupNet is to learn a multiscale hypergraph whose node is the agent and hyperedge is the interaction; and then, leverage this multiscale hypergraph to learn agent and interaction embedding.
\subsection{Multiscale hypergraph topology inference}
\label{subsection:topology}
\vspace{-1mm}
To comprehensively model group-wise interactions at multiple scales, we consider inferring a multiscale hypergraph from agent dynamics to reflect interactions at various group sizes; see Figure \ref{fig:ms-nmp}. Mathematically, let $\mathcal{V} = \{ v_1, v_2, \cdots, v_N \}$ be a set of agents and $\mathcal{G} = \{\mathcal{G}^{(0)}, \mathcal{G}^{(1)}, ... \mathcal{G}^{(S)} \}$ be a multiscale hypergraph showing the agent connections. At any scale $s$, $\mathcal{G}^{(s)}  =(\mathcal{V}, \mathcal{E}^{(s)})$ has the hyperedge set $\mathcal{E}^{(s)} = \{e^{(s)}_1, e^{(s)}_2, \cdots, e^{(s)}_{M_s} \}$ to represent group-wise relations with $M_s$ hyperedges, each of which links a number of agents to represent the common relations. A larger $s$ indicates a larger scale of agent groups. Notably, $\mathcal{G}^{(0)} =(\mathcal{V},\mathcal{E}^{(0)})$ is a specific hypergraph whose edges model the finest pair-wise agent connections. The topology of each $\mathcal{G}^{(s)}$ can be represented as an incidence matrix $\mathbf{H}^{(s)} \in \R^{|\mathcal{V}|\times |\mathcal{E}^{(s)}|}$ where $\mathbf{H}^{(s)}_{i, j}=1$ if the $i$th node is included in the $j$th hyperedge, otherwise $\mathbf{H}^{(s)}_{i, j}=0$.


\textbf{Affinity modeling.} The trajectory interactions are in a stealth mode and nontrivial to capture since there are no explicitly predefined graph topology for relation description. To infer a multiscale hypergraph, we construct hyperedges by grouping agents that have highly correlated trajectories, whose correlations could be measured by mapping the trajectories as a high-dimensional feature vector. Concretely, the trajectory embedding of the $i$th agent is $\mathbf{q}_i = f_{\mathrm{Q}}({\X^-_i}) \in \R^{d},$ where $f_{\mathrm{Q}}(\cdot)$ can be a MLP. We then compute an affinity matrix $\mathbf{A} \in \R^{N\times N}$ to reflect the correlation of any two agents. The $(i,j)$th element of $\mathbf{A}$ is:
\begin{equation}
\label{eq:attention}
  \setlength{\abovedisplayskip}{2pt}
  \setlength{\belowdisplayskip}{2pt}
    \mathbf{A}_{i,j} \ = \  \mathbf{q}_i^{\top} \mathbf{q}_j / (\|\mathbf{q}_i\|_2 \|\mathbf{q}_j\|_2),
\end{equation}
which represents the relational weight of the $i$th agent and the $j$th agent reflecting the correlation between two agents. We normalize the embeddings to ensure the trajectory representations to carry unit energy for stable relation estimation.

\textbf{Hyperedge forming.} Given the affinity matrix, we then form hyperedges at various scales. At the $0$th scale, we consider pair-wise connections. Each node connects the nodes that have the largest affinity scores with it, leading to $M_0$ edges in the incident matrix $\mathbf{H}^{(0)}$. For the other scales, we consider group-wise connections. Intuitively, agents in a group should have high correlation with each other. We thus find the groups of nodes by looking for high-density submatrices in the affinity matrix $\mathbf{A}$. We first assign a sequence of increasing group sizes $\{K^{(s)}\}_{s=1}^S$; and then, for each node, we find a highly correlated group at any scale $s$, leading to $K^{(s)}$ groups/hyperedges in each scale. 
Mathematically, let the $i$th hyperedge $e^{(s)}_i$ at the $s$th scale be the one associated with the $i$th node $v_i$. This hyperedge is obtained by solving the following optimization problem of searching a $K^{(s)}\times K^{(s)}$ submatrix:
\begin{equation}
   \setlength{\abovedisplayskip}{2pt}
   \setlength{\belowdisplayskip}{1pt}
\label{eq:opt}
   \begin{aligned}
   & e^{(s)}_i = \arg \mathop{\operatorname{max}}\limits_{\Omega \subseteq \mathcal{V}} \left \| \mathbf{A}_{\Omega, \Omega}\right\|_{\mathrm 1,\mathrm 1}, \\ 
   & \mathrm{s.t.}\;|\Omega|=K^{(s)}, ~v_i \in \Omega, ~ i=1,\dots,N,
   \end{aligned}
\end{equation}
where the entrywise matrix norm $\|\cdot\|_{\mathrm 1,\mathrm 1}$ denotes the sum of the absolute values of all elements. The objective finds the most correlated agents and links them together to consider the group behaviors. The first constraint limits the group size and the second requires the participance of the $i$th node. In this way, we could form at least one hyperedge that belongs to each node at each scale. To solve the optimization~\eqref{eq:opt}, when the total number of agents is small, we could adopt an enumeration algorithm to search for the optimum solution; otherwise, we employ a greedy algorithm approximation that first selects $v_i$ and then adds new nodes of maximum affinity value with $v_i$ at each move sequentially.

The incidence matrices for a multiscale hypergraph are thus $\{\mathbf{H}^{(0)} \in \R ^{N\times NK^{(0)}}, \{ \mathbf{H}^{(s)} \in \R ^{N\times N} \}_{s=1}^S\}$, where $\mathbf{H}^{(0)}$ includes $NK^{(0)}$ edges to consider the pair-wise interactions and $\mathbf{H}^{(s)} (s \geq 1)$ includes $N$ hyperedges to reflect the group-wise interactions within $K^{(s)}$ agents. Different from the common multiscale graph whose node numbers vary from scales \cite{li2020dynamic,gao2019graph}, our multiscale hypergraph has fixed node numbers, yet different hyperedge sizes on various scales. 
Note that all the edges and hyperedges are selected based on the same affinity matrix. This design brings two benefits: i) it is computationally efficient to search for high-order relationships from a single matrix; and ii) it makes the training of the affinity matrix more stable and informative through back-propagation.

Compared to~\cite{kipf2018neural,hu2020collaborative,li2020evolvegraph}, our topology inference method is novel from two aspects. First, our method actively infers a graph structure through learning; while many previous methods directly adopt a fixed fully-connected topology. Second, our method models both pair-wise and group-wise connections at multiple scales; while the previous methods only model the pair-wise connections at a single scale.

\subsection{Multiscale hypergraph neural message passing}
\label{subsection:nmp}
\vspace{-1mm}
To learn the patterns of agents trajectories given the inferred multiscale hypergraph, we customize a multiscale hypergraph neural message passing method obtaining the embeddings of agents and interactions iteratively through node-to-hyperedge and hyperedge-to-node, see Figure \ref{fig:ms-nmp}.
Specifically, we first initialize the agent embedding from trajectory of each agent; that is, at any scale, for the $i$th agent, $v_i$, its initial embedding is $\mathbf{v}_{i} = \mathbf{q}_i \in \mathbb{R}^d$ (see Sec.~\ref{subsection:topology}). At each scale, in the node-to-hyperedge phase, groups of agent embeddings are aggregated to get the interaction embeddings. In the hyperedge-to-node phase, each agent embedding is updated according to associated interaction embeddings. We execute the node-to-hyperedge and hyperedge-to-node for several iterations. We finally obtain the representation of each agent by fusing its embeddings across all the scales.

\textbf{Node-to-hyperedge phase.} To promote a representation for relational reasoning, we exhibit an interaction embedding with three elements: \textit{neural interaction strength}, representing the intensity of the interaction, \textit{neural interaction category}, reflecting agents interaction category, and \textit{per-category function}, modeling how interaction process of this category works. At each scale of hypergraph, for the $i$th hyperedge, $e_{i}$, its interaction embedding is obtained through
\begin{equation}
\label{eq:decouple}
   \setlength{\abovedisplayskip}{1pt}
   \setlength{\belowdisplayskip}{1pt}
    \mathbf{e}_{i} = r_{i}  \sum\limits_{\ell=1}^L  c_{i,\ell} \mathcal{F}_{\ell} \Big(\sum\limits_{v_j \in e_{i}} \mathbf{v}_{j} \Big) \in \R^{d}, 
\end{equation}
where $r_{i}$ is its neural interaction strength. $c_{i,\ell}$, the $\ell$th element of a category vector ${\bf c}_{i} \in [0,1]^{L}$, denotes the probability of the $\ell$th neural interaction category within $L$ possible categories. For each category, we assign a learnable category function $\mathcal{F}_{\ell}(\cdot)$ implemented by MLPs. Each of these three elements is trainable in an end-to-end framework.

To obtain the neural interaction strength $r_{i}$, and neural interaction category ${\bf c}_{i}$, we leverage a collective embedding that is a hidden state to reflect the overall information of agents in a group. For hypergraph at any scale, the collective embedding of $e_{i}$ is obtained by the weighted sum of all the agent embedding associated with the hyperedge $e_{i}$; that is,
$\setlength{\abovedisplayskip}{4pt}
   \setlength{\belowdisplayskip}{4pt}
    \mathbf{z}_{i} = \sum\limits_{v_j\in e_{i}} w_{j} \mathbf{v}_{j}$, 
where $w_{j} = \mathcal{F}_{\rm w} \Big( \mathbf{v}_{j}, \sum\limits_{v_m \in e_{i}} \mathbf{v}_{m} \Big),$
with $\mathcal{F}_{\rm w}(\cdot)$ implemented by an MLP. The weight $w_{j}$ reflects the contribution of the $j$th node to the $i$th group. We then use the collective embedding to infer the neural interaction strength, $r_{i}$, and the neural interaction category, ${\bf c}_{i}$; that is,
\begin{equation*}
   \setlength{\abovedisplayskip}{2pt}
   \setlength{\belowdisplayskip}{2pt}
r_{i} = \sigma\left(\mathcal{F}_{\rm r}(\mathbf{z}_{i})\right),
~~
\mathbf{c}_{i} = \mathrm{softmax}\left(( \mathcal{F}_{\rm c}( \mathbf{z}_{i})+\mathbf{g})/{\tau}\right),
\end{equation*}
where $\sigma(\cdot)$ is a sigmoid function to constrain the strength values, $\mathbf{g}$ is a vector whose elements are i.i.d. sampled from $\mathrm{Gumbel}(0,1)$ distribution and $\tau$ is the temperature controlling the smoothness of type distribution. We use the Gumbel softmax to make a continuous approximation of the discrete distribution following~\cite{maddison2016concrete}. $\mathcal{F}_{\rm r}(\cdot)$ and  $\mathcal{F}_{\rm c}(\cdot)$ are modeled by MLPs. The inference of the neural interaction strength, neural interaction category and per-category function considers the group behavior through the shared collective embedding.

\begin{figure*}[t] 
\centering
\includegraphics[width=0.9\textwidth]{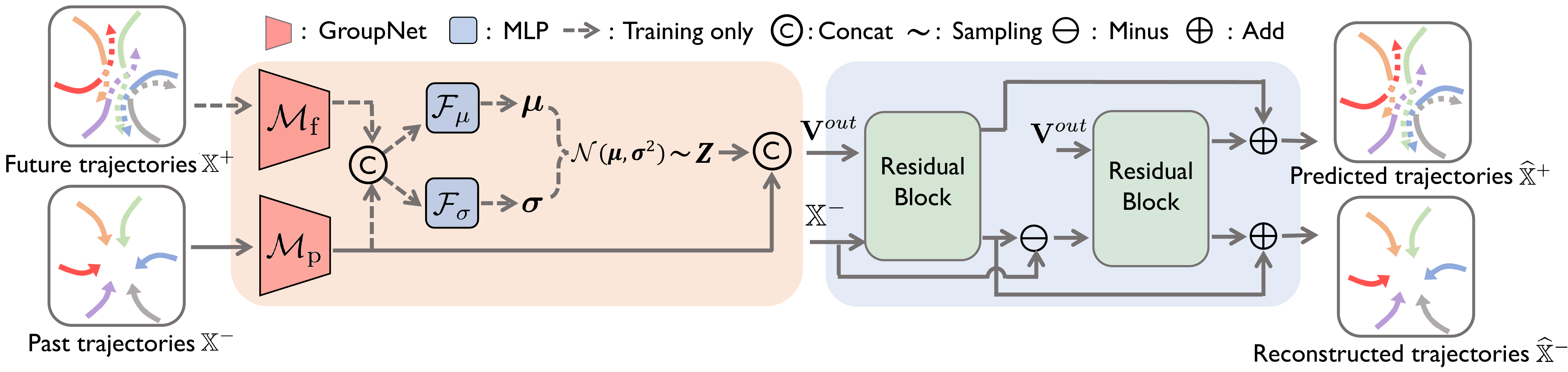}
\vspace{-3mm}
\caption{\small The architecture of the trajectory prediction system. The system consists of two processes: the encoding process (orange part) and the decoding process (blue part). Encoding process obtains the hidden representation for each agent. Decoding process aims to both predict the future trajectory and reconstruct the past trajectory given the hidden representation and past trajectory of each agent.}
\label{fig:system}
\vspace{-15pt}
\end{figure*}

\textbf{Hyperedge-to-node phase.} Given the interaction embeddings learnt from groups of agents, we then update each agent embedding by considering all the associated interactions. For any scale of hypergraph, let $\mathcal{E}_i = \{e_{j}|v_{i} \in e_{j}\}$ be the set of hyperedges associated with the $i$th node $v_i$.  The embedding of the $i$th agent in the next iteration is updated as 
\begin{equation*}
     \setlength{\abovedisplayskip}{2pt}
   \setlength{\belowdisplayskip}{1pt}
\mathbf{v}_{i} \leftarrow f_{\rm v} \big(\big[\mathbf{v}_{i}, \sum\limits_{e_j \in \mathcal{E}_i} \mathbf{e}_{j} \big]\big) \in \mathbb{R}^d,
\end{equation*}
where $f_{\rm v}(\cdot)$ is a trainable MLP; $[\cdot ,\cdot]$ denotes the embedding concatenation of one node and the associated hyperedges to absorb the influence from interactions to an agent.

At any scale $s$, we repeat the node-to-hyperedge and hyperedge-to-node phases for several times and obtain the embedding of the $i$th node, $\mathbf{v}_{i}^{(s)}$.
In this way, we compute the embeddings of the $i$th node, $\mathbf{v}_{i}^{(0)}, \mathbf{v}_{i}^{(1)} \dots, \mathbf{v}_{i}^{(S)}$, across $S$ scales in parallel. Note that the trainable layers in various scales are NOT shared. We finally concatenate the agent embedding across all the scales to obtain the comprehensive agents' embedding, formulated as $\mathbf{v}_{i} = [\mathbf{v}^{(0)}_{i}, \mathbf{v}^{(1)}_{i}, \cdots, \mathbf{v}^{(S)}_{i}] \in \mathbb{R}^{d(S+1)}$.

Compared to previous methods NRI \cite{kipf2018neural}, dNRI~\cite{graber2020dynamic}, and EvolveGraph \cite{li2020evolvegraph}, which achieve relational reasoning using neural message passing, the proposed hypergraph neural message passing is novel from three aspects. First, previous \cite{kipf2018neural,graber2020dynamic,li2020evolvegraph} adopt neural message passing on ordinary graphs; while we consider neural message passing on a series of hypergraphs, promoting more comprehensive information propagation and aggregation. Second, previous \cite{kipf2018neural,graber2020dynamic,li2020evolvegraph} only infer the interaction category; while our method can infer both interaction category and strength through a three-element representation format. Third, the interaction category inferred by previous \cite{kipf2018neural,graber2020dynamic,li2020evolvegraph} is the final output after message passing; while the interaction category and strength inferred by our method are the intermediate features that involve in neural message passing. Our design not only promotes relational reasoning in neural message passing, but also improves the prediction performance.

\vspace{-2mm}
\section{Prediction System with GroupNet}
\vspace{-1mm}
We apply GroupNet into a basic multi-agent trajectory prediction system based on the CVAE framework \cite{sohn2015learning} to handle the stochasticity of each agent's behavior. Let $\log p(\mathbb{X}^{+} \mid \mathbb{X}^{-})$ be the log-likelihood of future trajectories $\mathbb{X}^{+}$ conditioned on the past trajectories $\mathbb{X}^{-}$. The corresponding evidence lower bound (ELBO) is:
\begin{equation*}
   \setlength{\abovedisplayskip}{4pt}
   \setlength{\belowdisplayskip}{2pt}
\begin{aligned}
    \operatorname{log}p(\mathbb{X}^{+} \mid \mathbb{X}^{-}) \ge & \;\mathbb{E}_{q(\mathbf{Z}\mid \mathbb{X}^{+},\mathbb{X}^{-})}\operatorname{log}p(\mathbb{X}^{+}\mid \mathbf{Z},\mathbb{X}^{-}) \\
    &- \operatorname{KL}(q(\mathbf{Z}\mid \mathbb{X}^{+}
    ,\mathbb{X}^{-}) \| p(\mathbf{Z}\mid \mathbb{X}^{-})),
\end{aligned}
\end{equation*}
where $\mathbf{Z} \in \R^{N\times d_z}$ is the latent codes corresponding to all the agents; $p(\mathbf{Z} | \mathbb{X}^{-})$ is the conditional prior of $\mathbf{Z}$, which is set to be a Gaussian distribution.
Here we implement $q(\mathbf{Z} | \mathbb{X}^{+},\mathbb{X}^{-})$ by an encoding process for embedding learning, and implement $p(\mathbb{X}^{+}| \mathbf{Z},\mathbb{X}^{-})$ by a decoding process that predicts the future $\mathbb{X}^{+}$. The system is sketched in Figure~\ref{fig:system}. 

\textbf{Encoding process.}
The encoding process aims to generate the Gaussian parameters of the approximate posterior distribution. Mathematically, the Gaussian parameters generation of the approximate posterior distribution and the conditional prior distribution is given by:
\begin{equation*}
  \setlength{\abovedisplayskip}{1pt}
  \setlength{\belowdisplayskip}{1pt}
\begin{aligned}
    &\mathbf{V}^{+} \ = \ \mathcal{M}_\mathrm{f}( \mathbb{X}^+ ), \quad \mathbf{V}^{-} = \mathcal{M}_\mathrm{p}(\mathbb{X}^-),\\
    &\bm{\mu}_{q}  = \mathcal{F}_{\mathrm{\mu}}([\mathbf{V}^{+}, \mathbf{V}^{-}]), \quad \bm{\sigma}_{q}  = \mathcal{F}_{\mathrm{\sigma}}([\mathbf{V}^{+}, \mathbf{V}^{-}]),
\end{aligned}
\end{equation*}
where $\mathcal{M}_\mathrm{f}(\cdot)$ and $\mathcal{M}_\mathrm{p}(\cdot)$ is aforementioned GroupNet to get future agent embedding $\mathbf{V}^{+}$ and past agent embedding $\mathbf{V}^{-}$. $\bm{\mu}_{q}$, $\bm{\sigma}_{q}$ are the mean and variation of the approximate posterior distribution. $\mathcal{F}_{\mathrm{\mu}}(\cdot)$, $\mathcal{F}_{\mathrm{\sigma}}(\cdot)$ are MLPs. We sample latent code of possible future trajectories $\mathbf{Z}  \sim  \mathcal{N}(\bm{\mu}_{q},\operatorname{Diag}(\bm{\sigma}_{q}^2))$. In the testing phase, we sample $\mathbf{Z}$ from the prior distribution $\mathcal{N}(0,\lambda \mathbf{I})$, where $\lambda$ is a hyperparameter. We concatenate the latent code $\mathbf{Z}$ with the embedding of agent past trajectories $\mathbf{V}^\mathrm{-}$ as the output of encoding process: $\mathbf{V}^{\mathrm{out}} = [\mathbf{Z},\mathbf{V}^\mathrm{-}]$. 

\textbf{Decoding process.} We apply a residual decoder similar with \cite{cao2021spectral} which takes both the output embedding of the encoding process and past trajectories as input. The residual decoder aims to both predict the future trajectories and reconstruct the past trajectories, which avoids input information loss. The residual decoder consists of two decoding blocks with the same structure.  Mathematically, the calculation of the two blocks $\mathcal{F}_{Block1}(\cdot)$, $\mathcal{F}_{Block2}(\cdot)$ are:
\begin{equation*}
  \setlength{\abovedisplayskip}{1pt}
  \setlength{\belowdisplayskip}{1pt}
\begin{aligned}
    &\widehat{\mathbb{X}}^+_1, \widehat{\mathbb{X}}^-_1 = \mathcal{F}_{Block1}(\mathbf{V}^{out},\mathbb{X}^-),\\
    &\widehat{\mathbb{X}}^+_2, \widehat{\mathbb{X}}^-_2 = \mathcal{F}_{Block2}(\mathbf{V}^{out},\mathbb{X}^- - \widehat{\mathbb{X}}^-_1 ),
\end{aligned}
\end{equation*}
where $\widehat{\mathbb{X}}^+_1$, $\widehat{\mathbb{X}}^+_2$ and $\widehat{\mathbb{X}}^-_1$, $\widehat{\mathbb{X}}^-_2$ are predicted future trajectories and reconstructed past trajectories of two blocks. Each block consists of a GRU encoder to encode the sequence and two MLPs as the output header. The final prediction of future trajectories $\widehat{\mathbb{X}}^+$ and reconstruction of past trajectories $\widehat{\mathbb{X}}^-$ is the sum of two decoding blocks:
\begin{equation*}
  \setlength{\abovedisplayskip}{3pt}
  \setlength{\belowdisplayskip}{0pt}
\begin{aligned}
\widehat{\mathbb{X}}^+ = \widehat{\mathbb{X}}^+_1 + \widehat{\mathbb{X}}^+_2,\quad
\widehat{\mathbb{X}}^- = \widehat{\mathbb{X}}^-_1 + \widehat{\mathbb{X}}^-_2.
\end{aligned}
\end{equation*}


\textbf{Loss function.}
To train the prediction system, we minimize an overall loss function formed by an evidence lower bound loss, a past trajectory reconstruction loss and a variety loss in Social-GAN \cite{gupta2018social} to optimize the “best” prediction:
\begin{equation*}
   \setlength{\abovedisplayskip}{2pt}
   \setlength{\belowdisplayskip}{2pt}
\begin{aligned}
    &\mathcal{L}_{\mathrm{elbo}} =   
     \alpha \|\widehat{\mathbb{X}}^{+}-\mathbb{X}^{+}\|_{2}^{2} + \beta \operatorname{KL}(\mathcal{N}(\bm{\mu}_{q}, \operatorname{Diag}(\bm{\sigma}_{q}^2) \|\mathcal{N}(0,\lambda \mathbf{I})),\\
    &\mathcal{L}_{\mathrm{rec}} =  \gamma \|\widehat{\mathbb{X}}^{-}-\mathbb{X}^{-}\|_{2}^{2}, \mathcal{L}_{\mathrm{variety}} = \mathop{min}\limits_{k} \|\widehat{\mathbb{X}}^{+(k)}-\mathbb{X}^{+}\|_{2}^{2},\\
    &\mathcal{L} = \mathcal{L}_{\mathrm{elbo}} + \mathcal{L}_{\mathrm{rec}} + \mathcal{L}_{\mathrm{variety}},
\end{aligned}
\end{equation*}
where $\alpha$, $\beta$ and $\gamma$ are hyperparameters to control the weights, $\|\cdot\|_{2}$ denotes the $\ell_2$ norm and $\operatorname{KL}(\cdot||\cdot)$ denotes the KL divergence. In the variety loss, we predict multiple times and $\widehat{\mathbb{X}}^{+(k)}$ denotes $k$th prediction results.

\section{Experiments and Analysis}
\vspace{-1mm}
\subsection{Experimental setup}
\vspace{-2mm}
\textbf{Datasets.} 
We evaluate our method on both physical simulation datasets and three real-world public datasets: the NBA SportVU Dataset (NBA), the ETH-UCY Dataset~\cite{pellegrini2009you,lerner2007crowds}, and the Stanford Drone Dataset (SDD)~\cite{robicquet2016learning}. For the physical simulation datasets, since the ground truth of the interaction relationship can be accessed, we evaluate the ability of relational reasoning of our GroupNet. For the real-world datasets we evaluate the effectiveness of our proposed GroupNet. More details of datasets are introduced in Appendix.

\textbf{Metrics.}
Following ~\cite{li2020evolvegraph,alahi2016social,lee2017desire}, we use $\mathrm{minADE}_K$ and $\mathrm{minFDE}_K$ as evaluation metrics. $\mathrm{minADE}_K$ is the minimum among $K$ average distances of the $K$ predicted trajectories to the ground-truths in terms of the whole trajectories; $\mathrm{minFDE}_K$ shows the minimum distance among $K$ predicted endpoints to the ground-truth endpoints.



\textbf{Implementation details.}
In our model, all the MLPs have 3 layers with the ReLU activation function. The latent code dimension $d_z$ is 32. In the loss function, we set $\alpha,\beta,\gamma$ to 1.0 and the variance of the prior distribution $\lambda=1.0$. We consider all other agent at the $0$th scale. For NBA dataset, we set the scale to 2,5,11 players. For ETH dataset, we set the group scales to the half and the full of agents number and for SDD dataset to full of agent number in less than 100 pixels distance. To train the model, we apply Adam optimizer \cite{kingma2014adam} with learning rate $10^{-4}$ and decay every 10 epochs. See more details in Appendix. 

\vspace{-1mm}
\subsection{Validation on relational reasoning}
\label{sec:relation}
\begin{figure}[!t]
\centering
\subfloat[]{
\begin{minipage}[t]{0.33\linewidth}
\centering
\includegraphics[width=1\textwidth]{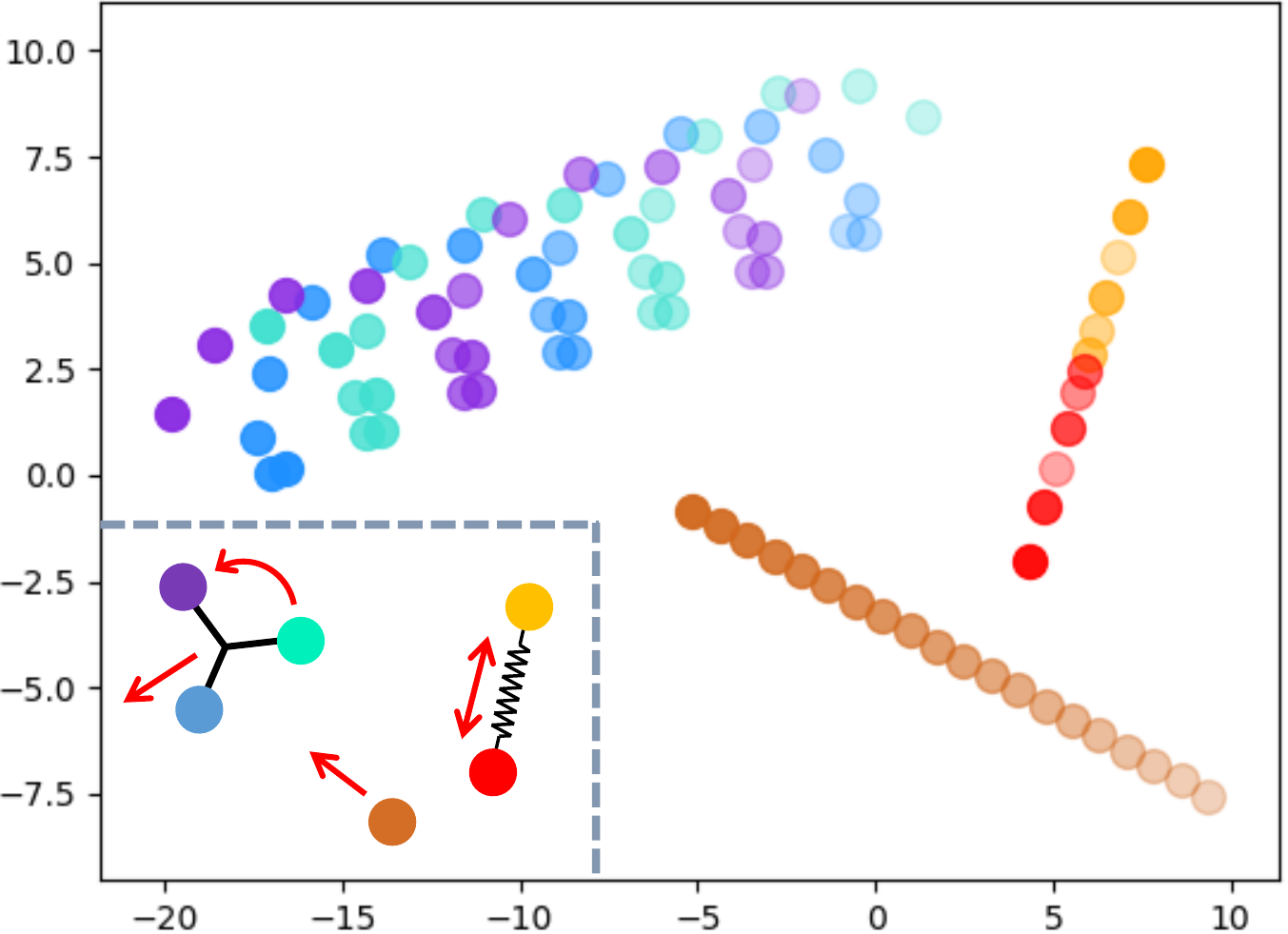}
\end{minipage}%
}%
\subfloat[]{
\begin{minipage}[t]{0.33\linewidth}
\centering
\includegraphics[width=1\textwidth]{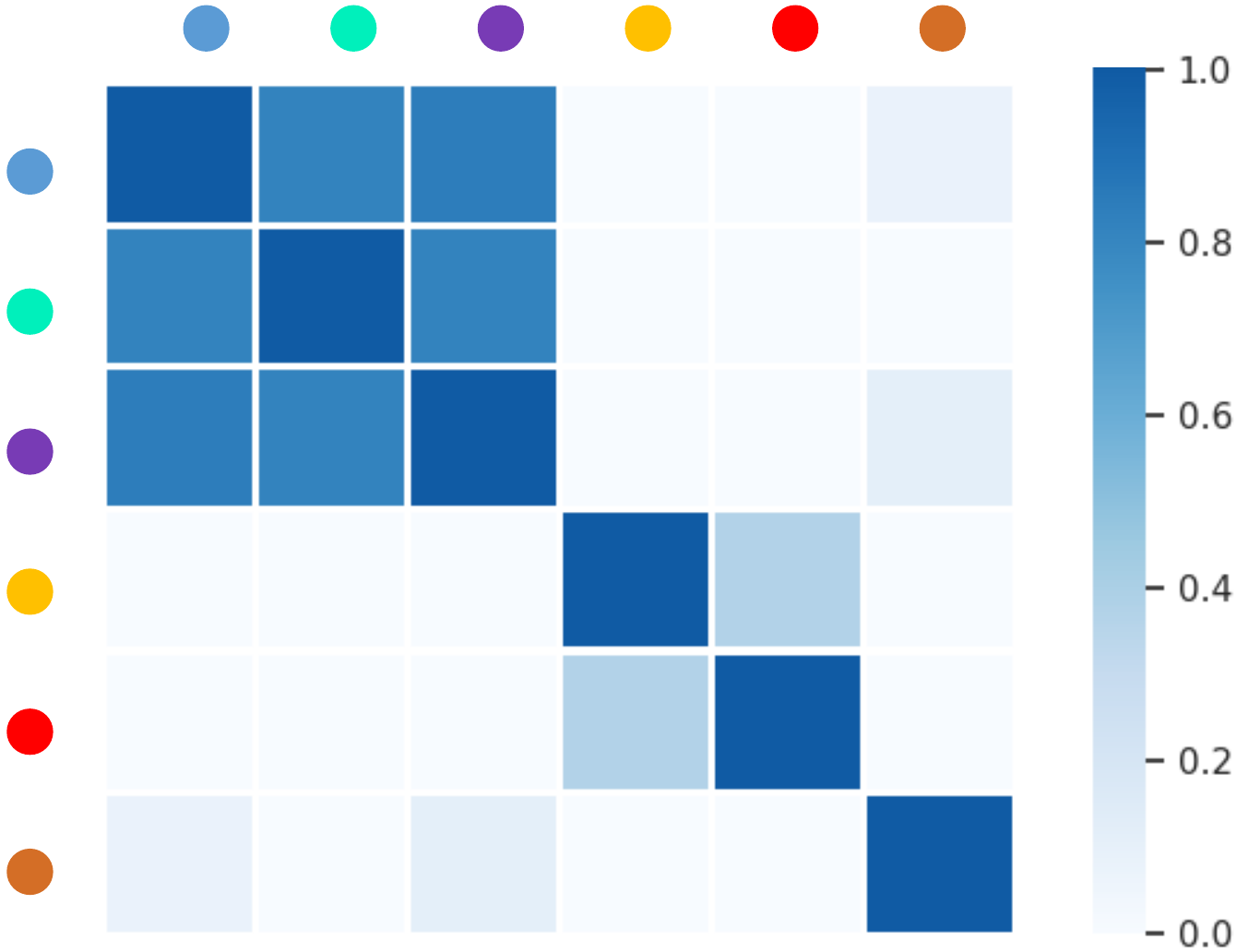}
\end{minipage}%
}%
\subfloat[]{
\begin{minipage}[t]{0.33\linewidth}
\centering
\includegraphics[width=1\textwidth]{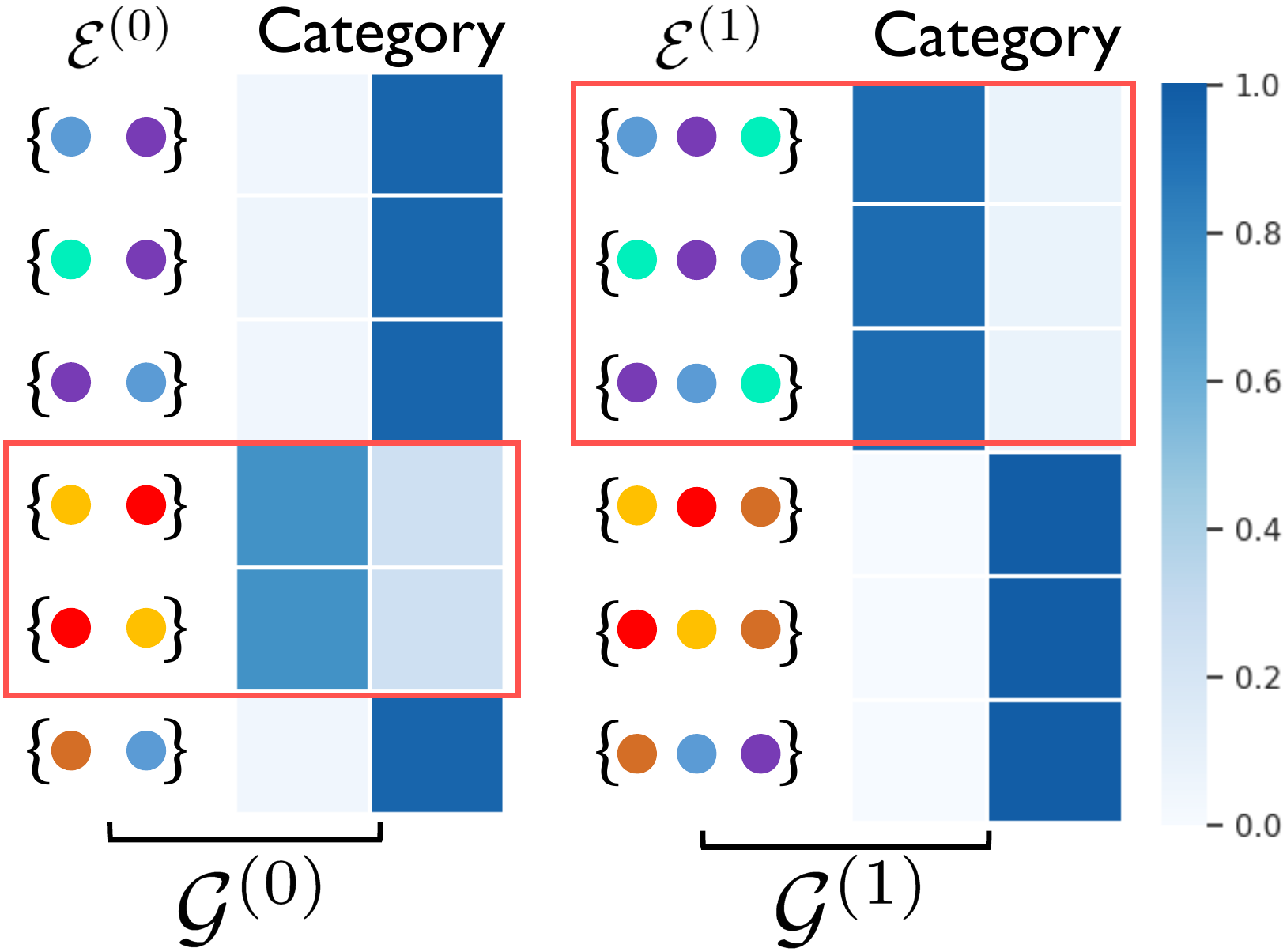}
\end{minipage}
}%
\centering
\vspace{-10pt}
\caption{\small Visualization of learnt group behavior. (a) The particle trajectories containing a three-group with a light bar, a two-group with a spring and an individual particle. (b) The heatmap of the learnt affinity matrix via affinity modeling. (c) The multiscale hypergraph topology via hyperedge forming and the interaction category vector of each hyperedge. The red box represents the three-group and two-group we inferred.}
\label{fig:heatmap}
\vspace{-10pt}
\end{figure}

\vspace{-1mm}
\mypar{Ability to capture group behaviors}
We consider six particles forming two scales of groups: the Y-shape light bar and the spring; see Figure \ref{fig:heatmap} (a) for an example. During the prediction process, GroupNet aims to find the group belongingness. Figure \ref{fig:heatmap} (b) shows the learnt affinity matrix in~\eqref{eq:attention}, where a darker color denotes higher affinity, which is revealed between the particles connected by the light bar or spring. 
Figure \ref{fig:heatmap} (c) illustrates the inferred (hyper)edges and their interaction categories (the category number is set to be 2). We see that i) the edge of spring and the hyperedge of light bar are inferred on our multiscale hypergraph $\{ \mathcal{G}^{(0)},\mathcal{G}^{(1)} \}$. ii) the spring and light bar have different interaction categories from other hyperedges, indicating that we effectively capture these two group interactions.

\begin{figure}[!t]
\centering
\subfloat[\small Free]{
\begin{minipage}[t]{0.33\linewidth}
\centering
\includegraphics[width=1\textwidth,height=0.73\textwidth]{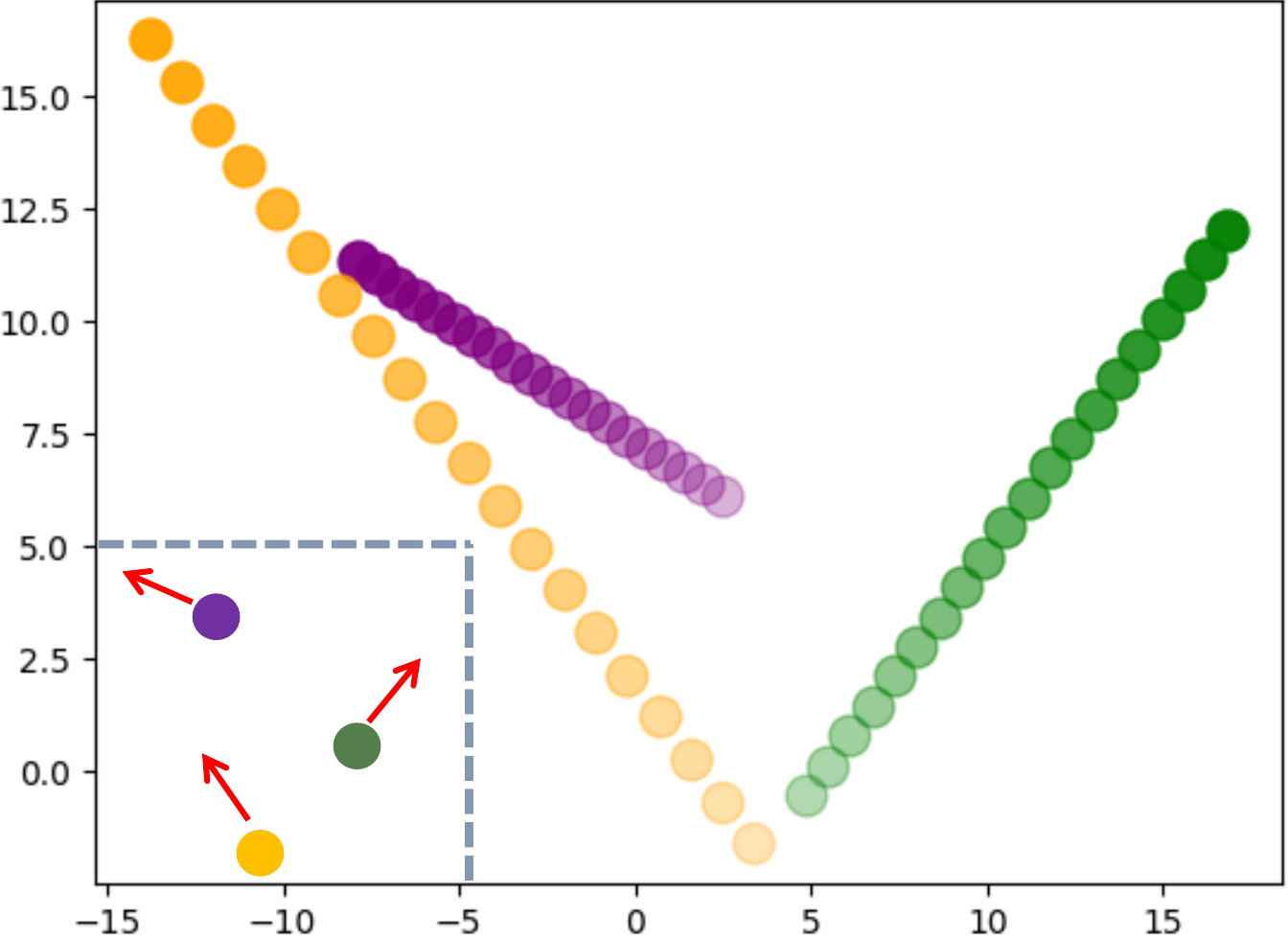}
\end{minipage}%
}%
\subfloat[\small Spring]{
\begin{minipage}[t]{0.33\linewidth}
\centering
\includegraphics[width=1\textwidth,height=0.73\textwidth]{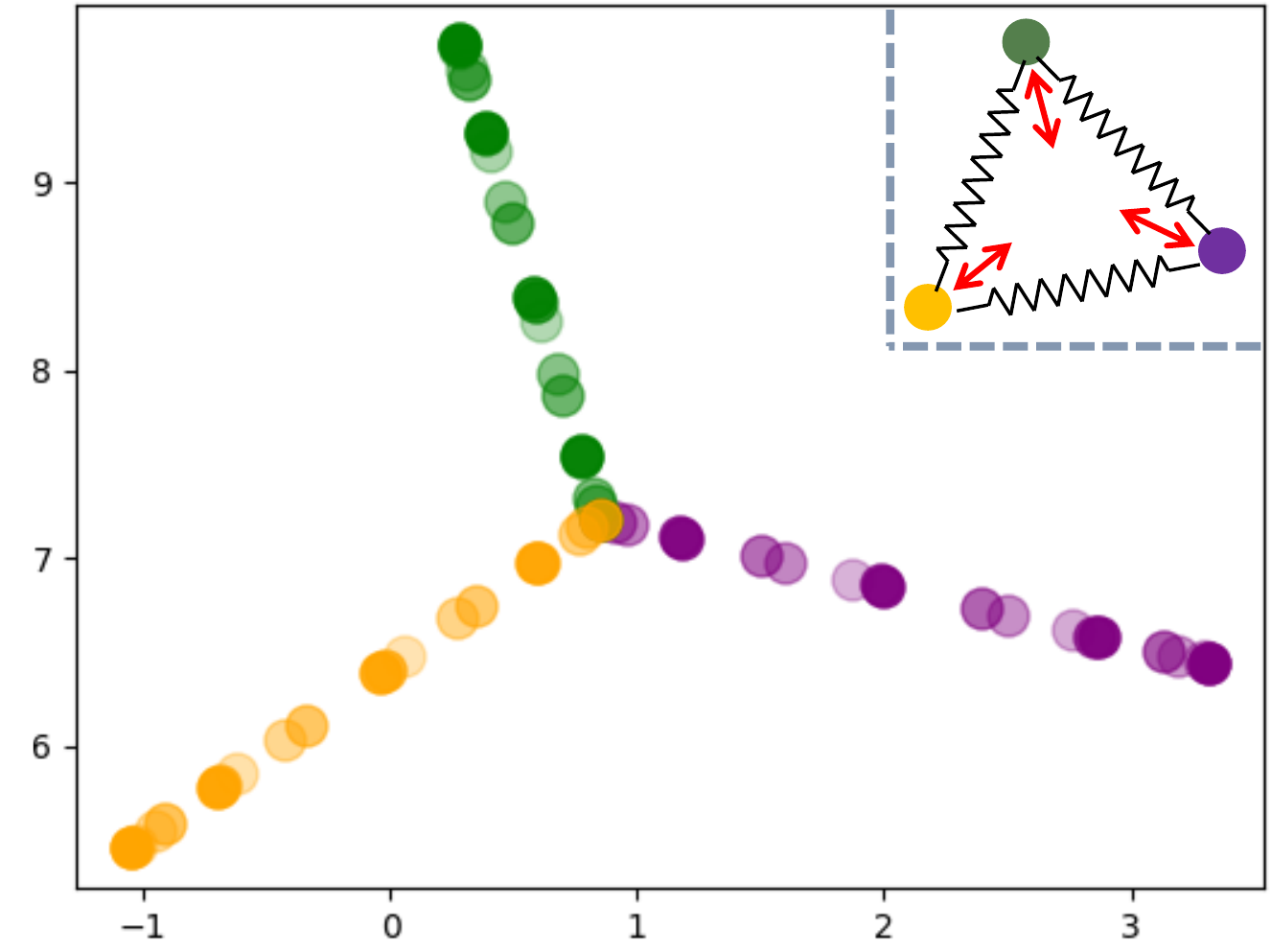}
\end{minipage}%
}%
\subfloat[\small Light bar]{
\begin{minipage}[t]{0.33\linewidth}
\centering
\includegraphics[width=1\textwidth,height=0.73\textwidth]{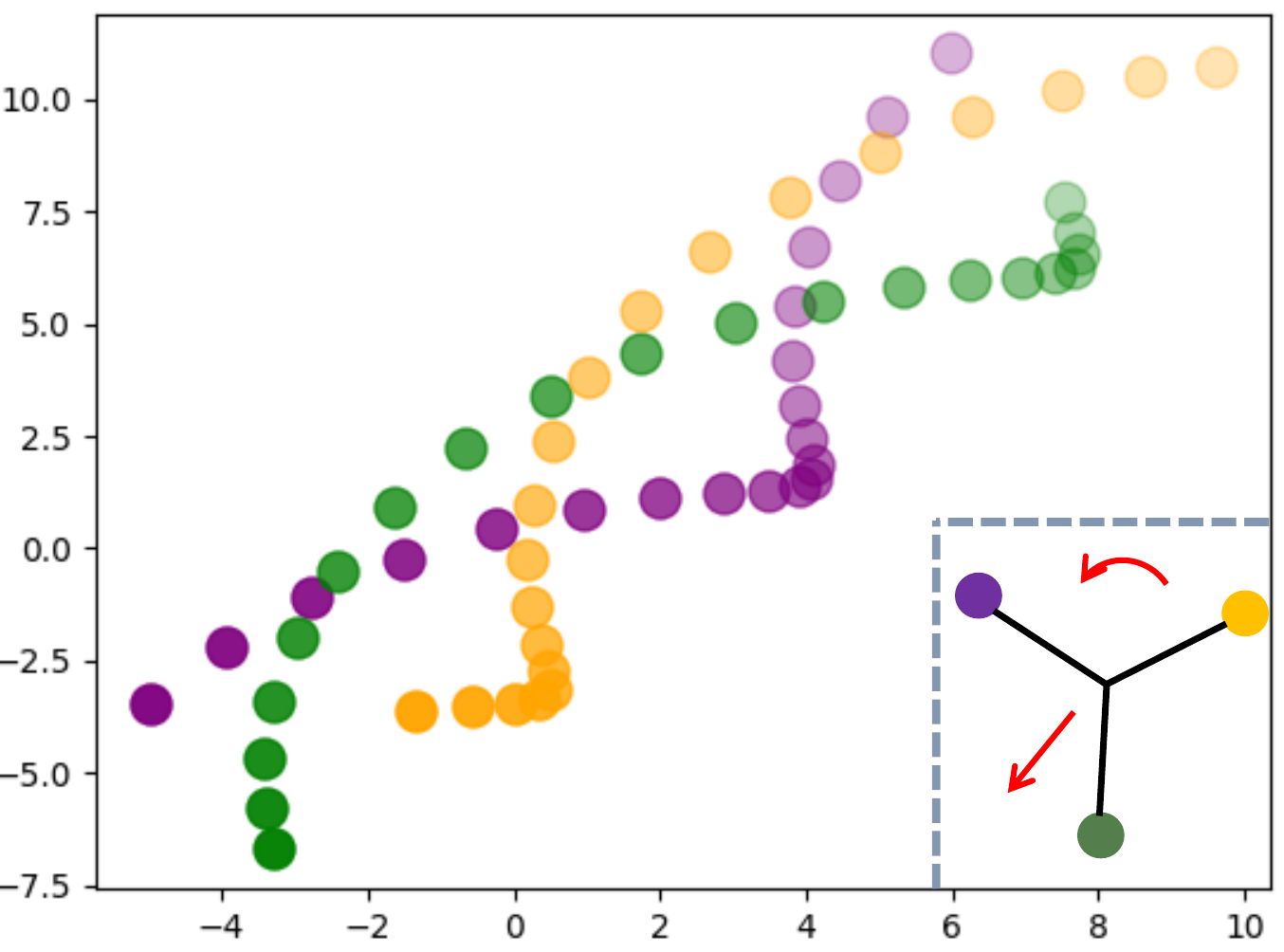}
\end{minipage}
}%
\centering
\vspace{-10pt}
\caption{Trajectory samples under three interaction categories.}
\label{fig:type}
\vspace{-10pt}
\end{figure}

\begin{table}[!t]
    \centering
    \caption{\small Comparison of accuracy (mean $\pm$ std in \%) of on interaction category recognition.}
    \small
    \vspace{-8pt}
    \setlength{\tabcolsep}{0.5mm}{
    \begin{tabular}{c|cccc|c}
        \hline
        \hline
        ~ &Corr.(path) &Corr.(LSTM) & NRI  & \textbf{Ours}  & Supervised \\
        \hline
        2-types &78.3$\pm$0.0&75.3$\pm$2.5 & 89.7$\pm$3.5 & \textbf{99.3}$\pm$0.3&99.5$\pm$0.2 \\
        3-types &\textbf{--}&\textbf{--} & 64.9$\pm$2.3 & \textbf{90.3}$\pm$0.9&98.7$\pm$0.7 \\
        \hline
        \hline
    \end{tabular}}
    \label{table:type}
    \vspace{-15pt}
\end{table}

\mypar{Ability to reason interaction category}
\label{sec:interaction_category} 
We consider three particles forming one group with three possible interaction categories: free, spring and light bar. Figure \ref{fig:type} gives examples of each interaction type. We compare the neural interaction category in~\eqref{eq:decouple} with the ground-truth category to achieve interaction recognition and report the recognition accuracy. This category recognition is unsupervised because the model only accesses the trajectories of particles without any prior information about the three types of interaction. We compare our method with three baselines~\cite{kipf2018neural} which consider only pair-wise interactions and an upper bound. To adapt them to group-wise actions, we use a majority vote for all pair-wise edges. The upper bound `Supervised' represents using the ground-truth category to train the neural interaction category. Table~\ref{table:type} reports the recognition accuracy on 2-type (only free and light bar type) and 3-type recognition tasks, which are averaged over 5 independent runs. `--' represents not applicable. We see that i) our method significantly outperforms three baselines, as we model the group interaction while NRI only considers pair-wise interaction; and ii) our method achieves similar performance to the upper bound, reflecting that our method is capable to reason the interaction category in an unsupervised manner. 


\begin{table*}[t]
\centering
\setlength{\tabcolsep}{1mm}\caption{minADE$_{20}$ / minFDE$_{20}$ (meters) of trajectory prediction (NBA dataset). $*$ denotes that NMMP is the previous best model on this dataset. \textbf{With GroupNet, NMMP works even much better. Simple CVAE with GroupNet achieves the current best result.}}
\vspace{-9pt}
\setlength{\tabcolsep}{2mm}{
\resizebox{\textwidth}{!}{
\begin{tabular}{l|ccccccc|cc|cc}
    \hline
    \hline
    Time
    & \makecell[c]{Social\\-LSTM\cite{alahi2016social}}
    &\makecell[c]{Social\\-GAN\cite{gupta2018social}} &\makecell[c]{Social-\\STGCNN\cite{mohamed2020social}} 
    &\makecell[c]{STGAT\\\cite{huang2019stgat}}
    & \makecell[c]{NRI\\\cite{kipf2018neural}}
    & \makecell[c]{STAR\\\cite{yu2020spatio}}
    & 
    \makecell[c]{PECNet\\\cite{mangalam2020not}} &
    \makecell[c]{NMMP*\\\cite{hu2020collaborative}} &
    \makecell[c]{GroupNet\\+NMMP}&\makecell[c]{CVAE}  &\makecell[c]{GroupNet\\+CVAE} 
    \\
\hline
     1.0s &0.45/0.67& 0.46/0.65&   0.36/0.50 &0.38/0.55 & 0.45/0.64 &0.43/0.65&0.51/0.76 &0.38/0.54 &0.38/0.54&0.37/0.52 &\textbf{0.34}/\textbf{0.48}  \\
     2.0s &0.88/1.53 &0.85/1.36& 0.75/0.99  & 0.73/1.18& 0.84/1.44& 0.77/1.28& 0.96/1.69&0.70/1.11&0.69/1.08&0.67/1.06&\textbf{0.62}/\textbf{0.95} \\
     3.0s & 1.33/2.38 &1.24/1.98&1.15/1.79&1.07/1.74&1.24/2.18 & 1.00/1.55& 1.41/2.52&1.01/1.61&0.98/1.47&0.96/1.51 &\textbf{0.87}/\textbf{1.31}  \\
     4.0s & 1.79/3.16&1.62/2.51& 1.59/2.37 & 1.41/2.22&1.62/2.84&1.26/2.04 &1.83/3.41&1.33/2.05 &1.25/1.80 &1.25/1.96 &\textbf{1.13}/\textbf{1.69}                     \\
    \hline
    \hline
\end{tabular}}
}
\label{table:nba}
\vspace{-9pt}
\end{table*}

\begin{table*}[!t]
\footnotesize
\centering
\setlength{\tabcolsep}{1mm}{\caption{minADE$_{20}$ / minFDE$_{20}$ (pixels) of trajectory prediction (SDD dataset). $*$ denotes that PECNet is the previous best model on this dataset. \textbf{With GroupNet, PECNet works even better. Simple CVAE with GroupNet achieves the current best ADE result.}}
\vspace{-9pt}
\resizebox{\textwidth}{!}{
\begin{tabular}{l|ccccccc|cc|cc}
\hline
\hline
    Time 
    &\makecell[c]{Social-\\LSTM\cite{alahi2016social}}
    & \makecell[c]{Social\\-GAN\cite{gupta2018social}}
    &\makecell[c]{SOPHIE\\\cite{sadeghian2019sophie}}  & \makecell[c]{Trajectron++\\\cite{salzmann2020trajectron++}}
    & \makecell[c]{NMMP\cite{hu2020collaborative}}
    & \makecell[c]{EvolveGraph\\\cite{li2020evolvegraph}}&CF-VAE \cite{bhattacharyya2019conditional} &\makecell[c]{PECNet*\\\cite{mangalam2020not}} &\makecell[c]{GroupNet\\+PECNet}& CVAE &\makecell[c]{GroupNet\\+CVAE}
    \\
\hline
     4.8s& 31.19/56.97 & 27.23/41.44
     & 16.27/29.38
     &19.30/32.70&14.67/26.72&13.90/22.90&12.60/22.30&9.96/15.88&9.65/\textbf{15.34}&10.22/18.18 &\textbf{9.31}/16.11 \\
\hline
\hline
\end{tabular}
\label{table:sdd}}}
\vspace{-9pt}
\end{table*}

\begin{table*}[!t]
\centering
\setlength{\tabcolsep}{1mm}\caption{minADE$_{20}$ / minFDE$_{20}$ (meters) of trajectory prediction (ETH-UCY dataset).  $*$ denotes that Trajectron++ is the previous best model on this dataset. \textbf{With GroupNet, Trajectron++ works even better. Simple CVAE with GroupNet achieves comparable results.}}
\vspace{-9pt}
\resizebox{\textwidth}{!}{
\begin{tabular}{l|cccccccc|cc|cc}
  \hline
  \hline
    Subset & \makecell[c]{Social-\\LSTM\cite{alahi2016social}}& \makecell[c]{Social-\\Attention\cite{vemula2018social}} &\makecell[c]{Social-\\GAN\cite{gupta2018social}}&\makecell[c]{SOPHIE\\\cite{sadeghian2019sophie}}&\makecell[c]{STGAT \cite{huang2019stgat}}&NMMP\cite{hu2020collaborative}&STAR\cite{yu2020spatio}&\makecell[c]{PECNet\\\cite{mangalam2020not}} &\makecell[c]{Trajectron++*\\\cite{salzmann2020trajectron++}} & \makecell[c]{GroupNet\\+Trajectron++}& CVAE& \makecell[c]{GroupNet\\+CVAE} \\
\hline
     ETH& 1.09/2.35&1.39/2.39&0.87/1.62&0.70/1.43&0.65/1.12&0.61/1.08 &\textbf{0.36}/\textbf{0.65}&0.54/0.87 &0.39/0.83&0.38/0.74&0.47/0.80&0.46/0.73 \\
     HOTEL&0.79/1.76&2.51/2.91&0.67/1.37&0.76/1.67&0.35/0.66&0.33/0.63&0.17/0.36&0.18/0.24&0.12/0.21&\textbf{0.11}/\textbf{0.20}&0.17/0.31&0.15/0.25\\
     UNIV&0.67/1.40&1.25/2.54&0.76/1.52&0.54/1.24&0.52/1.10&0.52/1.11&0.31/0.62&0.35/0.60&0.20/0.44&\textbf{0.19}/\textbf{0.40}&0.28/0.52&0.26/0.49  \\
     ZARA1&0.47/1.00&1.01/2.17&0.35/0.68&0.30/0.63&0.34/0.69&0.32/0.66&0.26/0.55&0.22/0.39 &0.15/0.33&\textbf{0.14}/\textbf{0.32}&0.25/0.49&0.21/0.39\\
     ZARA2&0.56/1.17&0.88/1.75&0.42/0.84&0.38/0.78&0.29/0.60&0.43/0.85&0.22/0.46&0.17/0.30&\textbf{0.11}/\textbf{0.25}&\textbf{0.11}/\textbf{0.25} &0.21/0.40&0.17/0.33\\
     AVG&  0.72/1.54&1.41/2.35&0.61/1.21&0.54/1.15&0.43/0.83&0.41/0.82&0.26/0.53&0.29/0.48&\textbf{0.19}/0.41&\textbf{0.19}/\textbf{0.38}&0.28/0.50&0.25/0.44\\
  \hline
  \hline
\end{tabular}}
\label{table:eth}
\vspace{-9pt}
\end{table*}

\begin{figure}[!t]
\centering
\subfloat[\small Low charge]{
\begin{minipage}[t]{0.33\linewidth}
\centering
\includegraphics[width=1\textwidth]{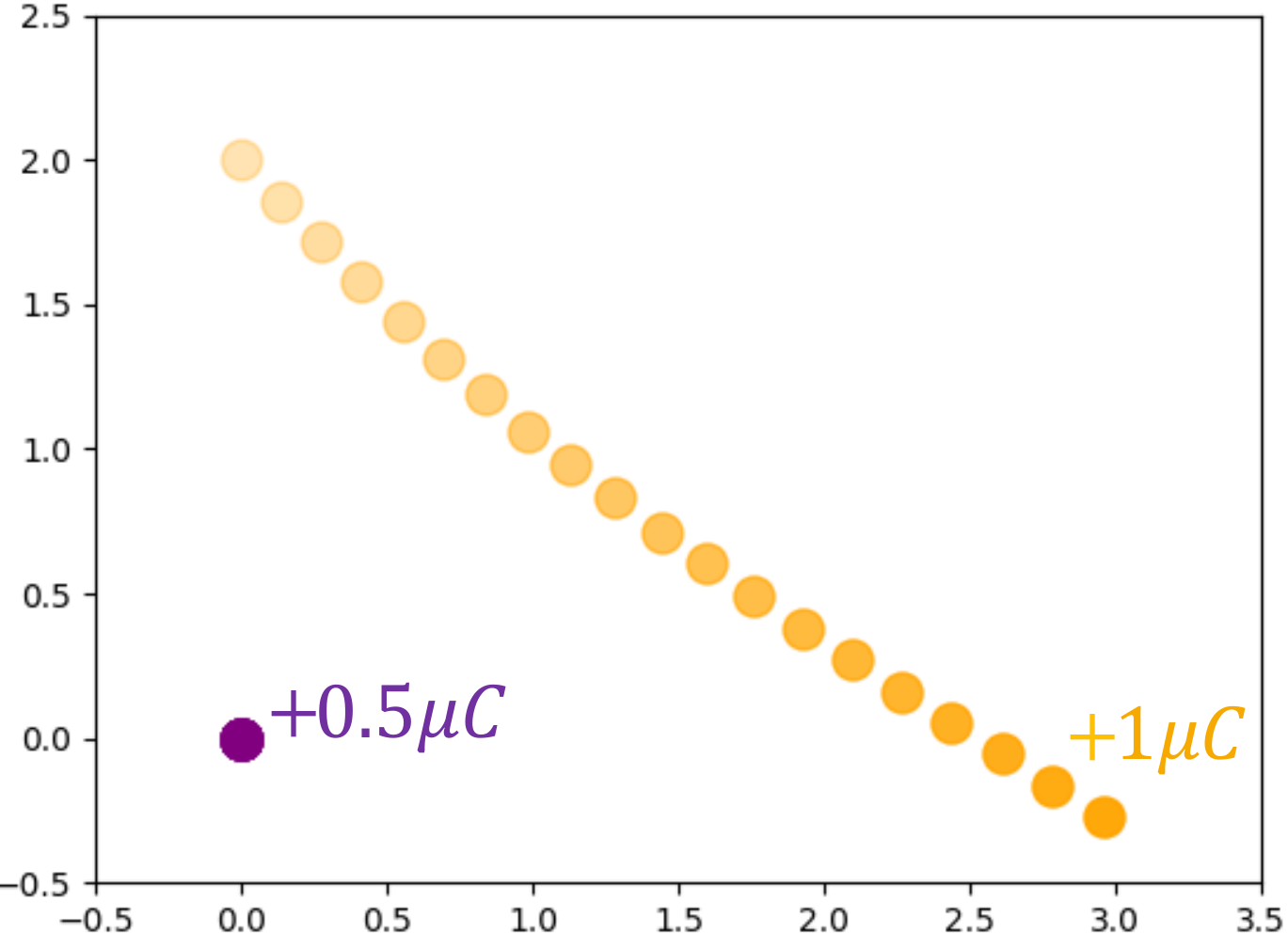}
\end{minipage}%
}%
\subfloat[\small High charge]{
\begin{minipage}[t]{0.33\linewidth}
\centering
\includegraphics[width=1\textwidth]{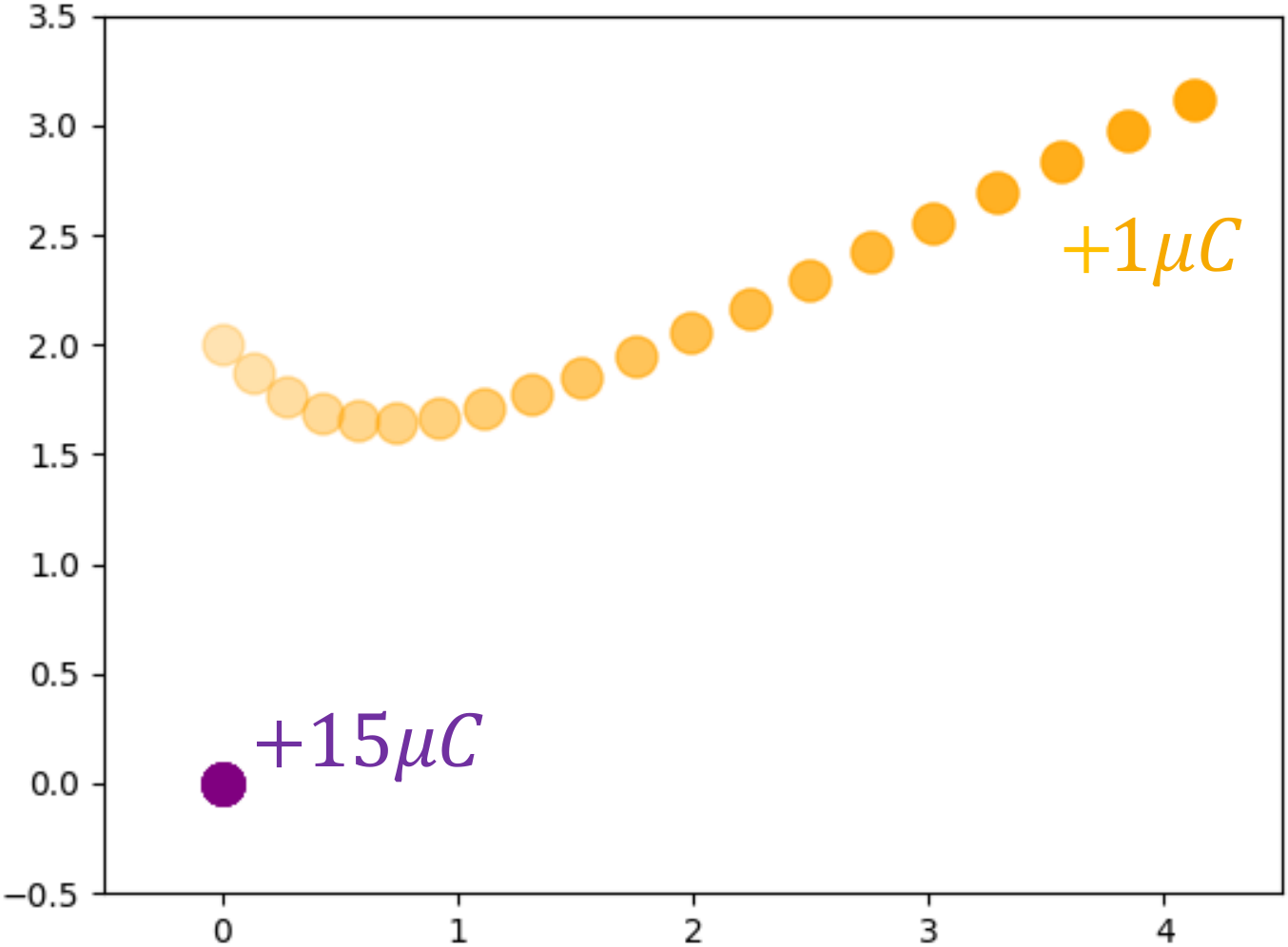}
\end{minipage}%
}%
\subfloat[\small Relation curve]{
\begin{minipage}[t]{0.33\linewidth}
\centering
\includegraphics[width=1\textwidth]{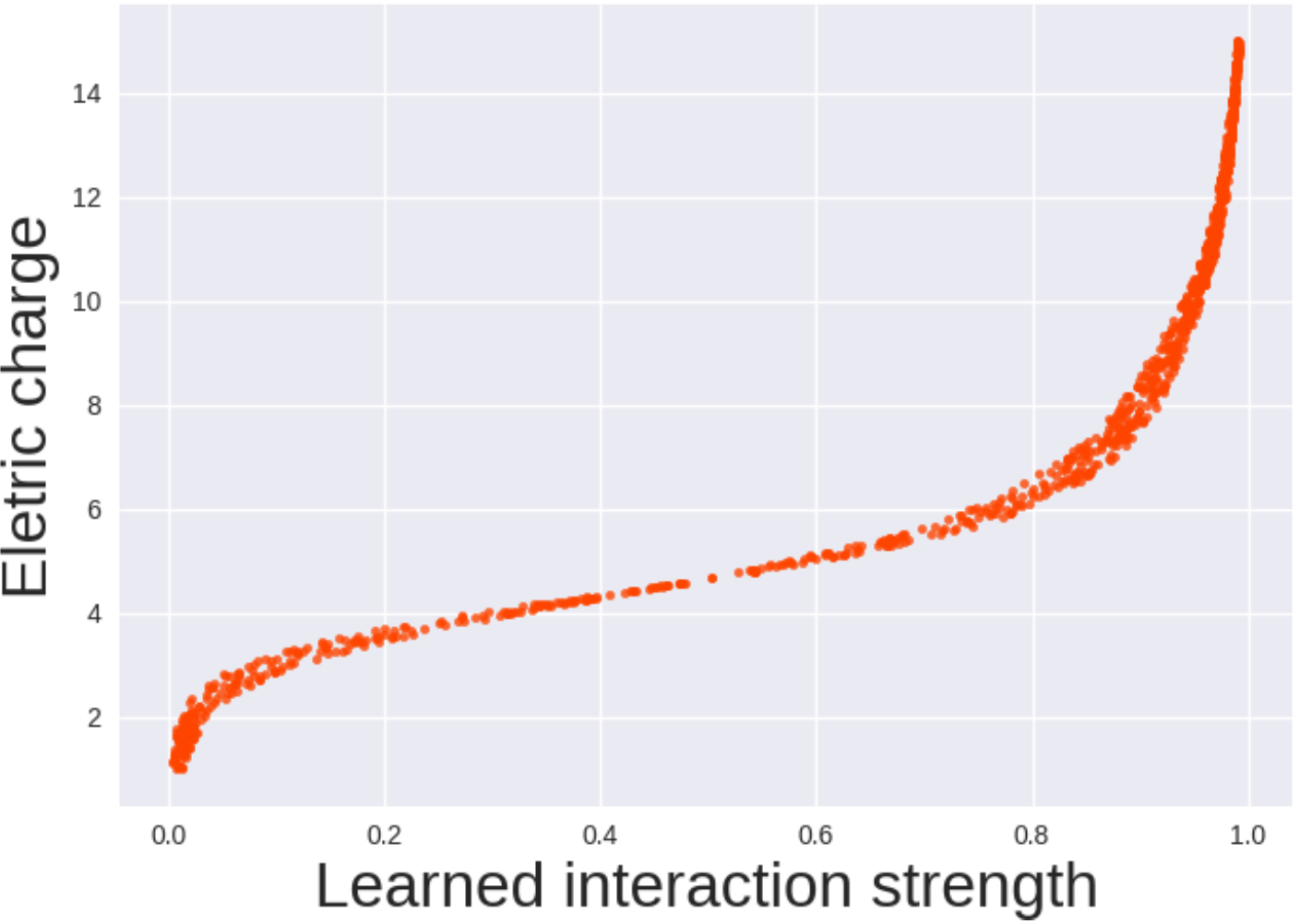}
\end{minipage}
}%
\centering
\vspace{-9pt}
\caption{\small Particles' trajectories and the curve of neural interaction strength with particle's electric charge.}
\label{fig:intensity}
\vspace{-12pt}
\end{figure}

\mypar{Ability to reason interaction strength}
We consider the movement of two charged particles which interact via Coulomb forces. Figure~\ref{fig:intensity} shows two examples of particle trajectories. A stronger repulsion the moving particle receives leading to a higher interaction strength. During the prediction we aim to validate whether the neural interaction strength in~\eqref{eq:decouple} can reflect the amount of charge without any direct supervision.
Figure~\ref{fig:intensity} (c) shows the relations between the learnt interaction strength ($x$-axis) and charge on the fixed particle ($y$-axis). We see that the neural interaction strength has a proportional relationship with the amount of charge, reflecting our model is capable to implicitly capture the interaction strength in an unsupervised manner. 

\vspace{-0.5mm}
\subsection{Validation on effectiveness}
\label{sec:results}
\vspace{-1mm}
\mypar{NBA dataset} We predict the future 10 timestamps (4.0s) based on the historical 5 timestamps (2.0s). Table \ref{table:nba} shows the comparison with nine state-of-the-art methods. We see that i) GroupNet with CVAE framework achieves a significant improvement over the state-of-the-art methods: the $\mathrm{minADE_{20}}$ and $\mathrm{minFDE_{20}}$ at 4.0s is reduced by $15.0\% / 17.6 \%$ comparing to the best baseline method (NMMP). The improvement over previous methods increases with timestamps, since the proposed multiscale hypergraph can capture more comprehensive interaction patterns and ii) We add GroupNet into previous the state-of-the-art method and replace its encoder (GroupNet+NMMP). Our GroupNet improve its performance by reducing $\mathrm{minADE_{20}}/\mathrm{minFDE_{20}}$ at 4.0s by $6.0\%/12.2\%$.

\mypar{SDD dataset}
Table~\ref{table:sdd} compares the proposed method with nine state-of-the-art methods. We see that i) GroupNet with CVAE framework achieves state-of-the-art performance and reduces the $\mathrm{minADE_{20}}$ by $6.5\%$ comparing to PECNet. ii) We add GroupNet into the previous best framework and replace its Social Pooling encoder (GroupNet+PECNet). Our GroupNet is capable to improve its performance by reducing $\mathrm{minADE_{20}}/\mathrm{minFDE_{20}}$ by $3.1\%/3.4\%$.

\mypar{ETH-UCY dataset}
Table~\ref{table:eth} compares the proposed method with nine state-of-the-art methods and the `AVG' means the averaged result over 5 subsets. We see that i) GroupNet with CVAE framework outperforms most of previous methods and achieves closing performance with the state-of-the-art method (Trajectron++); and ii) We add GroupNet into previous the state-of-the-art method and replace its encoder (GroupNet+Trajectron++). Our GroupNet is capable to improve its performance by reducing $\mathrm{minFDE_{20}}$ by $7.3\%$ and reaches a new state-of-the-art performance.

\begin{figure*}[!t]
\centering
\subfloat[\small NMMP]{
\begin{minipage}[t]{0.32\linewidth}
\centering
\includegraphics[width=1\textwidth]{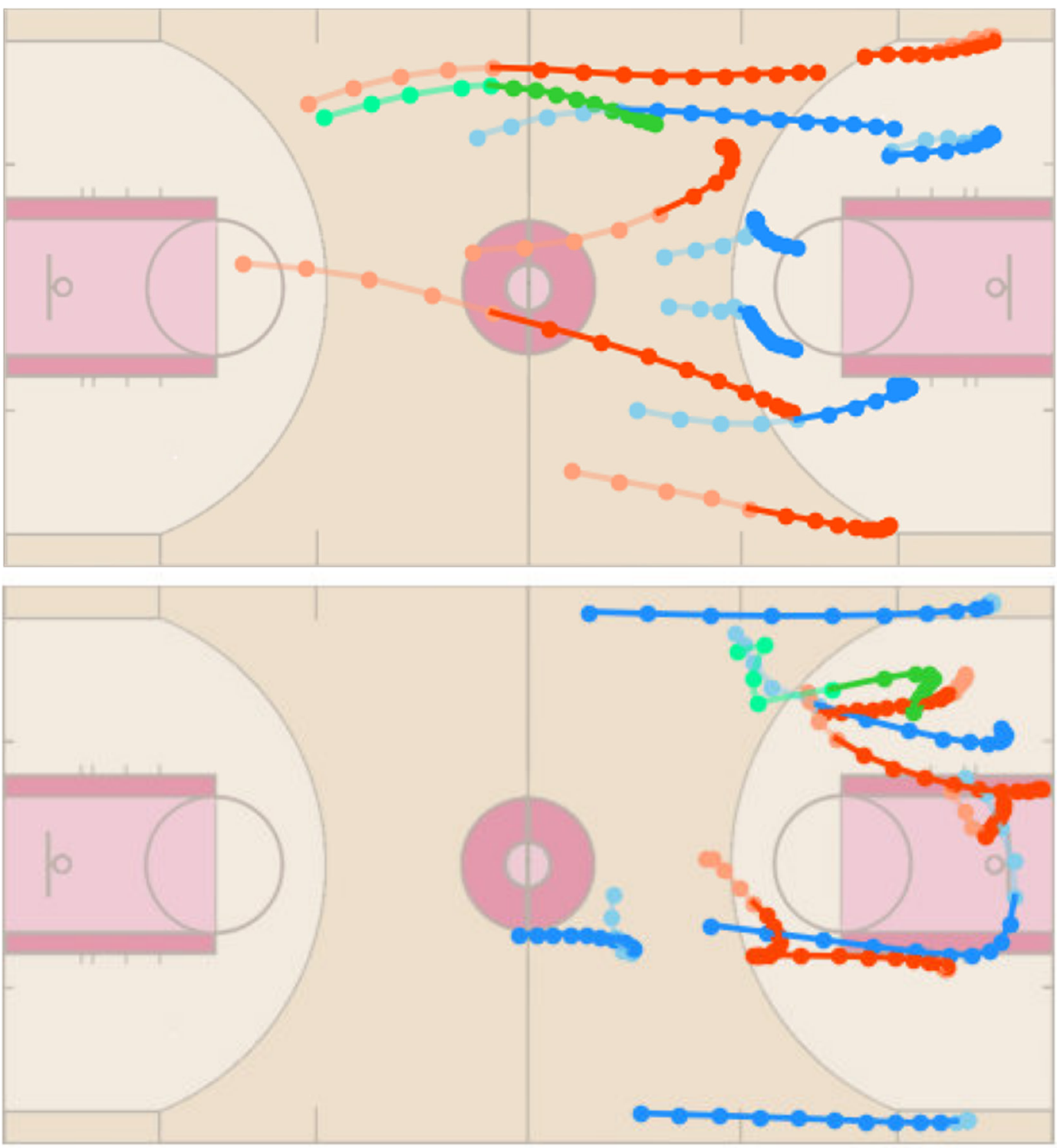}
\end{minipage}%
}%
\subfloat[\small  Ours]{
\begin{minipage}[t]{0.32\linewidth}
\centering
\includegraphics[width=1\textwidth]{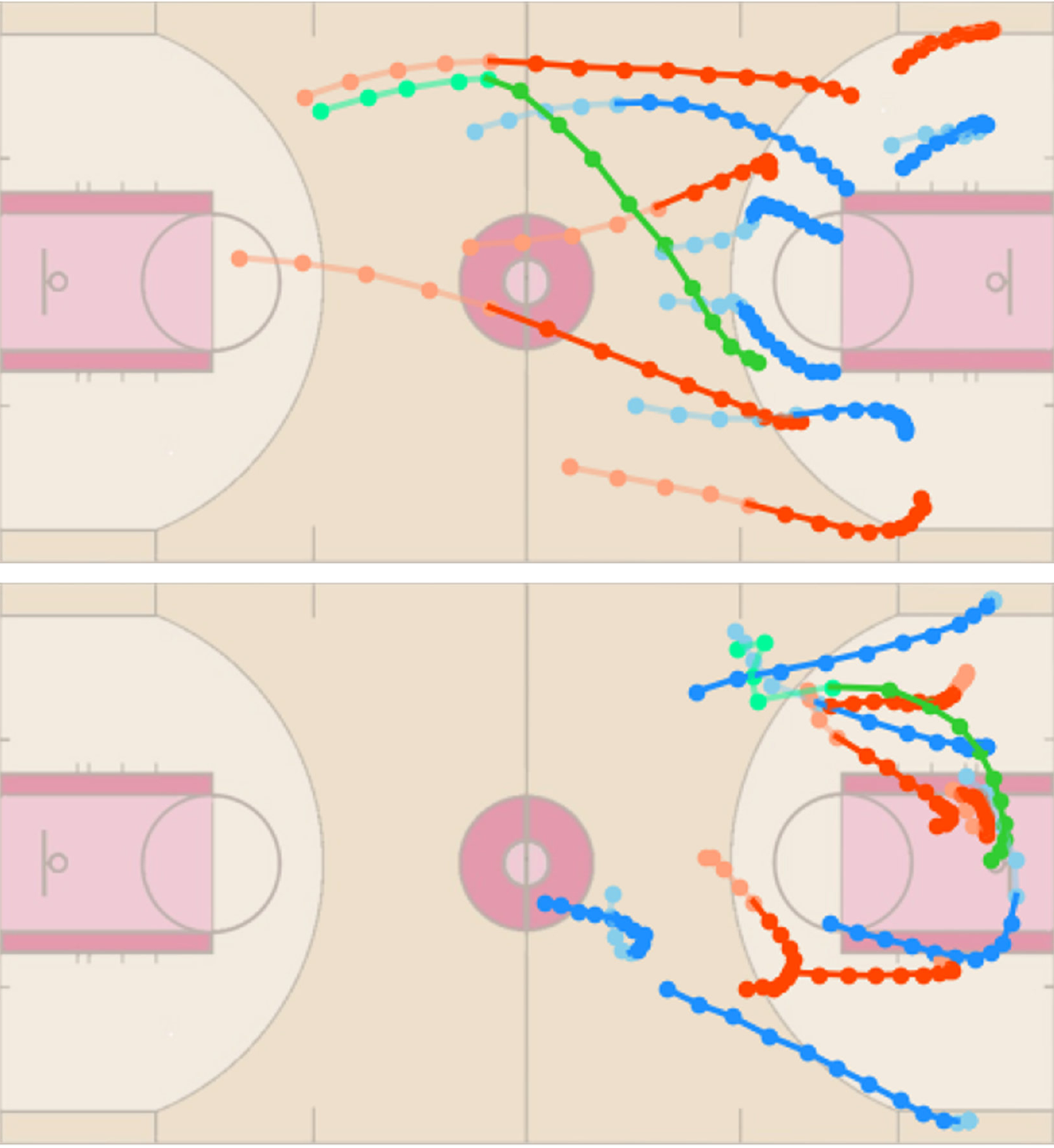}
\end{minipage}%
}%
\subfloat[\small GT]{
\begin{minipage}[t]{0.32\linewidth}
\centering
\includegraphics[width=1\textwidth]{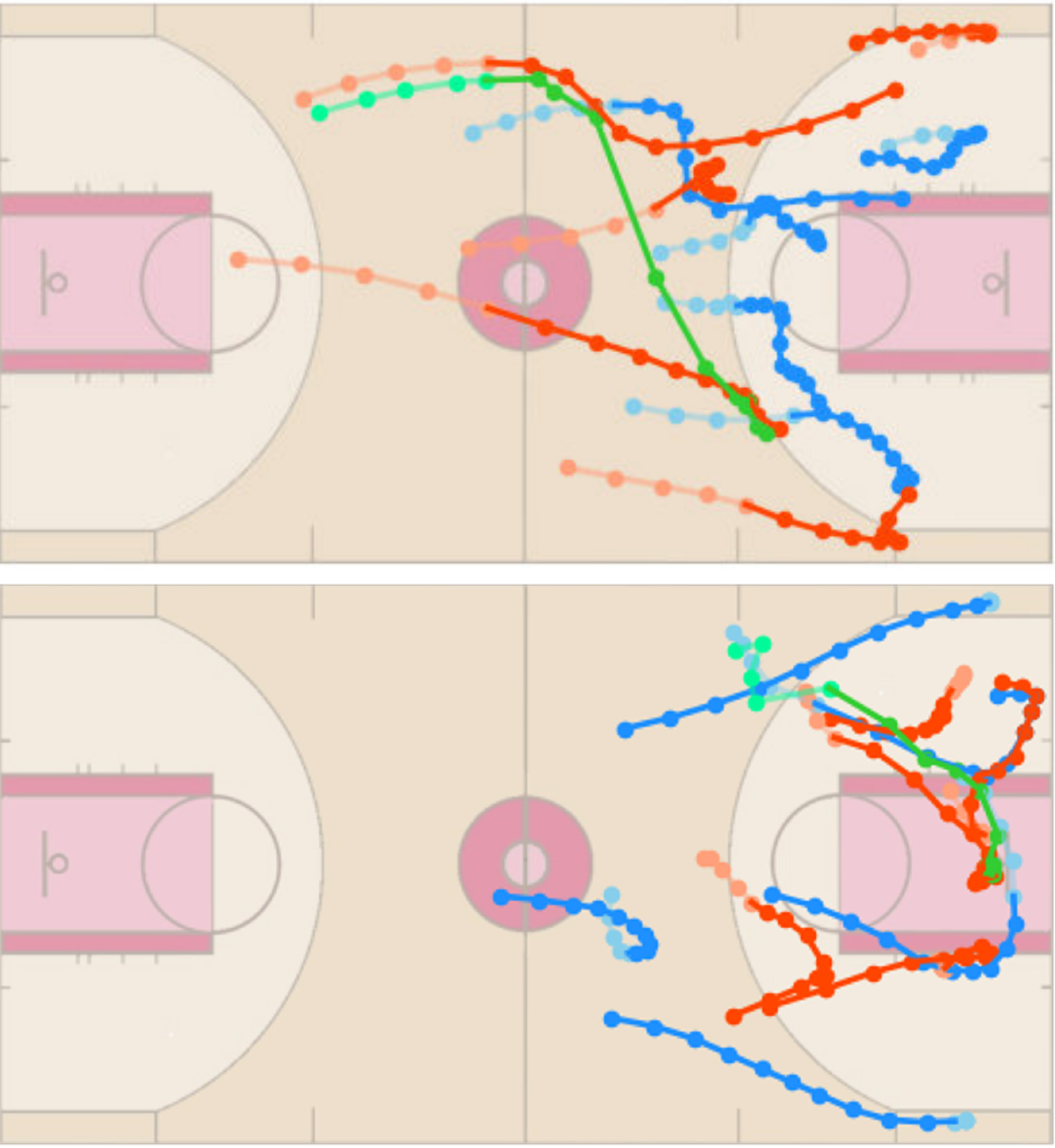}
\end{minipage}
}%
\centering
\vspace{-12pt}
\caption{\small Qualitative results on the NBA dataset. We plot the best trajectory among 20 predictions for the state-of-the-art method (NMMP), GroupNet with the CVAE framework (Ours) and ground truth (GT). The red/blue color represents players of two teams and the green color represents the basketball. Light color represents the past trajectory.}
\label{fig:result}
\vspace{-12pt}
\end{figure*}

\mypar{Discussion}
We see that previously, NMMP, PECNet and Trajctron++ are the best models on NBA, SDD and  ETH-UCY  datasets, respectively, and no single method achieves the best performances across all three datasets.~\emph{Here we analyze that this phenomenon may result from the inherence of the datasets.}  In SDD, there are relatively smooth trajectories and simple interactions among agents. Hence, once a specific destination is given, the predictions tend to be straightforward. This enables PECNet that emphasize accurate destinations prediction to achieve good results. Opposite to SDD, the NBA dataset entails much more complex agent interactions. Methods that pay more attention to interaction modeling would suit NBA setting well. For example, NMMP and STAR achieve leading performances on NBA. Lastly, ETH-UCY contains five subsets, each of which varies from each other. This matches Trajectron++ that incorporates multi-modal decoders to promote generalization.

Under this circumstance, our GroupNet shows commendable consistency: across all the datasets and methods, \emph{once GroupNet is plugged in to a prediction system, the performance can be significantly and consistently lifted up.} This reflects the robustness and generalization of our method. 

\begin{table}[!t]
    \caption{\small  Ablation studies of different feature extraction modules on the NBA dataset. We report minADE$_{20}$ / minFDE$_{20}$ (meters). }
    \vspace{-3mm}
    \centering
    \setlength{\tabcolsep}{1.5mm}{\small
\begin{tabular}{c|cccc}
\hline
\hline
\multicolumn{1}{c|}{Module} & \multicolumn{4}{c}{Prediction time}  \\ \hline
$\mathcal{M_\mathrm{p}}$ \& $\mathcal{M_\mathrm{f}}$ & 1.0s & 2.0s & 3.0s & 4.0s \\
\hline
MLP  & 0.37/0.52&0.67/1.06&0.96/1.51&1.25/1.96   \\
NMMP &0.35/0.49&0.63/0.97&0.90/1.36&1.17/1.76 \\
ATT     &  0.35/0.50    &  0.63/0.97   &   0.90/1.35   & 1.16/1.74  \\
GroupNet   &   \textbf{0.34}/\textbf{0.48}&\textbf{0.62}/\textbf{0.95}&\textbf{0.87}/\textbf{1.31}&\textbf{1.13}/\textbf{1.69} \\
         \hline
          \hline
\end{tabular}}
\label{table:sys}
\vspace{-3mm}
\end{table}

\mypar{Qualitative results}
Figure \ref{fig:result} compares the predicted trajectories of NMMP, GroupNet with the CVAE framework (Ours) and ground-truth (GT) trajectories on NBA dataset. We see that i) our method produces more precise predictions than previous state-of-the art method NMMP; and ii) for the scenes of fierce confrontation between two teams (the second row), our method has a larger improvement. This is because the proposed method captures more comprehensive interactions among players in complicated scenes.

\begin{table}[t]
    \caption{\small  minADE$_{20}$ / minFDE$_{20}$ (meters) of different group scales on NBA dataset.}
    \vspace{-3mm}
    \centering
    \small
    \setlength{\tabcolsep}{1mm}{
   \begin{tabular}{c|cccc}
    \hline
    \hline
         & \multicolumn{4}{c}{Prediction time} \\
    \hline
        Scale&1.0s&2.0s&3.0s&4.0s \\
    \hline

        1&0.37/0.52 & 0.67/1.05 & 0.97/1.50 &1.27/1.95  \\
        1,2& 0.35/0.49&0.63/0.97&0.89/1.35&1.16/1.75\\
        1,2,5&  0.35/0.49 & 0.62/0.95&0.88/1.32&1.14/1.71\\
        1,2,5,11& 0.34/0.48&0.62/0.95&0.87/1.31&1.13/1.69\\
        1,2,3,5,11& 0.34/0.48&0.62/0.94&0.87/1.31&1.13/1.70 \\
    \hline
    \hline
    \end{tabular}
    \label{table:scales}
  }
 \vspace{-4mm}
\end{table}

\vspace{-1mm}
\subsection{Ablation studies}
\vspace{-1mm}
\mypar{Effects of GroupNet in the system}
We perform extensive ablation studies on the NBA dataset to investigate the contribution of the key technical component GroupNet;see Table \ref{table:sys}. We apply three other feature extraction modules. The `MLP' means that we use a multi-layer perceptron. The `ATT' amd `NMMP' means that using the non-local attention mechanism in \cite{wang2018non} and using the graph neural message passing in \cite{hu2020collaborative} which models the pair-wise interactions between two agents. We see that the proposed GroupNet leads to superior performance comparing to the multi-layer perceptron, the attention mechanism and the graph-based mechanism.

\mypar{Effects of multiple scales}
Table \ref{table:scales} presents the effect of multiple scales in the GroupNet on the CVAE framework on the NBA dataset. We see that i) The performance increases at first when choosing more scales. ii) The performance becomes stable when having sufficient scales. 

\vspace{-1mm}
\section{Conclusion}
\vspace{-2mm}
This paper proposes GroupNet, which promotes a more comprehensive interaction modeling from two aspects: i) designing a multiscale hypergraph to capture both pair-wise and group-wise interactions; and ii) designing a three-element representation format of interaction embedding, including neural interaction strength, category and per-category function. We apply the proposed GroupNet as a key component to a CVAE-based prediction system and previous systems. Experiments on synthetic physics simulations and real-world datasets reveal the ability of relational reasoning and the effectiveness of GoupNet.

\mypar{Limitation and future work} In this work, the agent number is relatively small and the time window is relatively short due to the attribute of datasets. In the future, we will extend the network into more complex interacting systems and explore the long-term prediction task containing a time-varying number of agents with an adaptive GNN.
\section*{Acknowledgements}
\vspace{-2mm}
This research is partially supported by the National Key R\&D Program of China under Grant 2021ZD0112801, National Natural Science Foundation of China under Grant 62171276, the Science and Technology Commission of Shanghai Municipal under Grant 21511100900 and CCF-DiDi GAIA Research Collaboration Plan 202112. 


{\small
\bibliographystyle{ieee_fullname}
\bibliography{egbib}
}

\end{document}